\documentclass{article}
\usepackage{iclr2027_conference,times}

\usepackage{amsmath,amsfonts,bm}

\def\eqref#1{equation~\ref{#1}}

\def\1{\bm{1}}

\DeclareMathAlphabet{\mathsfit}{\encodingdefault}{\sfdefault}{m}{sl}
\SetMathAlphabet{\mathsfit}{bold}{\encodingdefault}{\sfdefault}{bx}{n}

\usepackage{graphicx}
\usepackage{float}
\usepackage{afterpage}
\usepackage{booktabs}
\usepackage{multirow}
\usepackage{subcaption}
\usepackage{amsmath}
\usepackage{amssymb}
\usepackage{xcolor}
\usepackage{url}
\usepackage{hyperref}
\definecolor{linkblue}{RGB}{0,60,140}
\hypersetup{colorlinks=true,linkcolor=linkblue,citecolor=linkblue,urlcolor=linkblue}
\usepackage[capitalise]{cleveref}

\title{SVGFaith: A Two-Layer Meta-Evaluation\\ of Faithfulness Metrics\\ for Text-to-SVG Generation}

\author{Guotao Liang$^{1,2}$\quad Haitao Zhou$^{2}$\quad Biao Liu$^{1}$\quad Huanxuan Liao$^{1}$\quad
Lingfeng Qiao$^{1}$\\
\textbf{Qian-Wen Zhang$^{1}$\thanks{Corresponding author.}\quad Qian Yu$^{2}$\quad Di Yin$^{1}$\quad
Xing Sun$^{1}$}\\
$^{1}$Tencent Youtu Lab\quad $^{2}$Visual Computing Group, Beijing}

\iclrfinalcopy
\makeatletter
\def\@maketitle{\vbox{\hsize\textwidth\centering
{\LARGE\sc \@title\par}
\begin{tabular}[t]{c}\bf\rule{\z@}{24pt}\@author\end{tabular}\par
\vskip 0.3in minus 0.1in}}
\makeatother
\renewcommand{\headrulewidth}{0pt}

\begin{document}

\maketitle

\begin{abstract}
Text-to-SVG systems are commonly evaluated by applying raster-image metrics to rasterised outputs, yet
whether these scores capture fine-grained prompt faithfulness remains unclear. We introduce
SVGFaith, a two-layer meta-evaluation framework in which the more faithful candidate of each pair is
determined by construction or by human judgment. SVGFaith-Diagnostic contains $1{,}842$ controlled
contrastive pairs across eight capability axes, split evenly between two directions: two renderings
compared against one prompt, and two descriptions compared against one rendering. SVGFaith-Arena
contains $1{,}327$ human-annotated comparisons of real generator outputs. On SVGFaith-Diagnostic,
CLIP and three other image--text embedding metrics reach at least $0.789$ pairwise accuracy on object
identity but at most $0.672$ on spatial relations and attribute binding, where chance is $0.5$. We
further introduce SVGFaith-Score, an interpretable pointwise metric. A domain-tuned decomposer turns
each prompt into a dependency graph of atomic visual requirements, a visual question answering model
scores whether each requirement is satisfied, and a requirement's score is discounted when its
prerequisites score low. In addition to the overall score, the per-requirement scores show which
requirements are judged unsatisfied and which inherit discounts from their prerequisites.
SVGFaith-Score significantly outperforms all nine baselines, including a Qwen3.5-122B-A10B judge, on
Arena and in both Diagnostic directions. Code and data: \url{https://github.com/Yukinonooo/SVGFaith}.
\end{abstract}

\section{Introduction}
\label{sec:intro}

\afterpage{%
\begin{figure}[t]
\centering
\includegraphics[width=\textwidth]{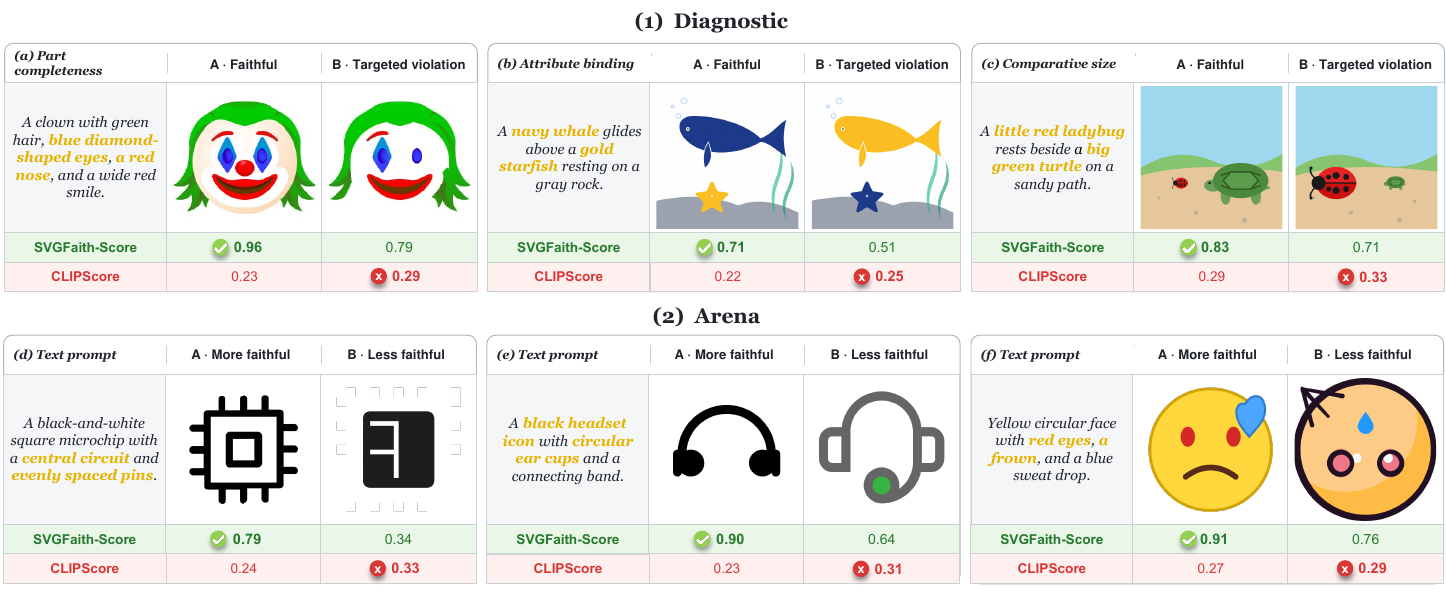}
\caption{Six faithfulness comparisons: three controlled SVGFaith-Diagnostic pairs
\textbf{(a--c)} and three human-annotated SVGFaith-Arena pairs \textbf{(d--f)}. In Diagnostic, the
left rendering satisfies the prompt and the right violates the highlighted constraint. In Arena,
the left rendering is judged more faithful and the right less faithful, although both may contain
errors. Here, \emph{CLIPScore} denotes the normalised image--text cosine from CLIP ViT-L/14
\citep{clip_radford_2021}, abbreviated as CLIP-L/14 elsewhere. In all six selected examples, it
ranks the right rendering higher, whereas SVGFaith-Score ranks the left higher.}
\label{fig:teaser}
\end{figure}%
}

Modern text-to-SVG methods produce renderable, visually coherent graphics, making prompt faithfulness a central evaluation question
\citep{omnisvg_yang_2025,wang2025internsvg,svgen_wang_2025,svgthinker_chen_2025, reasonsvg_xing_2025,renderintheloop_liang_2026}. In practice it is commonly
measured by an automatic score: a CLIP similarity between the prompt and the rasterised output \citep{clip_radford_2021}, or a rating from a vision--language judge \citep{gecko_wiles_2025,vectorgym_rodriguez_2026}. Both are widely used in raster text-to-image evaluation and have been adopted by vector benchmarks \citep{vgbench_zou_2024,svgenius_chen_2025,vectorgym_rodriguez_2026}, but aggregate generator rankings do not establish how reliably they resolve individual SVG constraints. A benchmark can therefore inherit a score's blind spots while still reporting a definitive ranking.

Meta-evaluation tests a metric on pairs whose more faithful candidate is determined by construction
or by human judgment, not by the metric itself
\citep{gecko_wiles_2025,t2iscorescore_saxon_2024,croc_leiter_2026}. In a constructed pair, one
candidate satisfies the prompt and the other violates a designated constraint, so a metric that
reverses the pair exposes a specific blind spot.
Such pairs cannot show whether the metric also works on real generator outputs, where several errors
of different severity may occur together. Human judgments of real outputs provide that test, but
when a metric disagrees with them, they do not reveal which difference the metric missed. SVGFaith
therefore contains one layer of each kind and reports the two separately.

The controlled layer, SVGFaith-Diagnostic, contains $1{,}842$ contrastive pairs over eight
capability axes (\Cref{fig:teaser}(a--c)). Its pairs are built by rule-based edits of the SVG
source, hard-negative retrieval or paired generation, and every retained pair meets the same human
review standard. Diagnostic is scored in two directions: $921$ pairs compare two renderings against
one prompt, and $921$ compare two descriptions against one rendering, so sensitivity to visual and
textual changes is probed separately. The real-output layer, SVGFaith-Arena, contains $1{,}327$
pairs of outputs from six generators (\Cref{fig:teaser}(d--f)). Each pair is judged by three
annotators under a rubric that asks for prompt faithfulness and excludes visual polish as a
criterion.

We also propose SVGFaith-Score, an interpretable pointwise metric. It decomposes the prompt into a
graph of atomic requirements, scores each requirement against one rendering with
satisfaction-aligned visual question answering (VQA) \citep{tifa_hu_2023,vqascore_lin_2024}, and
attenuates each fact by its prerequisites, so an attribute of an absent object is discounted. The
graph is built by SVGFaith-Decomposer, which we obtain by domain fine-tuning Qwen3.5-9B with
teacher-generated SFT, rejection-sampling SFT and Group Relative Policy Optimization (GRPO)
\citep{deepseekmath_shao_2024}; a separate Qwen3.5-9B serves as the VQA backend. Beyond the scalar
score, the node scores localise low-scoring requirements and trace inherited discounts through the
graph. Because each candidate is scored on its own, the metric can evaluate a single output or rank
any number of candidates. It can therefore also serve as a faithfulness verifier for filtering or
reranking generated samples, or as a reward signal for future text-to-SVG training.

We compare SVGFaith-Score with nine widely used evaluators on both layers, spanning image--text
embeddings, human-preference reward models and vision--language scorers (\Cref{sec:exp}).
SVGFaith-Score is the most accurate on Arena and in both Diagnostic directions, and its Arena
accuracy of $0.880$ closely matches the $0.875$ mean agreement of individual annotators with the
final labels. The baselines, in contrast, show failures that their aggregate accuracy hides.
When two renderings are compared against one prompt, CLIP-L/14 orders most object-identity pairs
correctly but fewer than half of the spatial-relation pairs (\Cref{fig:axis-chart}); more
generally, image--text embedding metrics recognise which objects appear far better than how they
are arranged or which attributes they carry. Reward models trained on human preferences fail on
other constraints: in the same setting, ImageReward \citep{imagereward_xu_2023}, one of the
strongest baselines on Arena, and HPS v2.1 \citep{wu2023humanv2} each order fewer than half of the
negation pairs correctly. Scaling a vision--language judge from Qwen3.5-9B to Qwen3.5-122B-A10B
\citep{qwen35_qwen_2026} improves accuracy on both layers, but the larger judge still trails
SVGFaith-Score (\Cref{tab:judge-scale}). Finally, all ten methods, including SVGFaith-Score,
become less accurate on Arena pairs in which a stronger generator's output is judged less faithful
(\Cref{sec:exp-slices}). This reliability gap is associated with generator identity, but these
pairs can also differ in semantic-error type, severity, opponent composition and annotation
difficulty, so its mechanism remains open. A preference for generator-associated visual style is
one possible explanation.

In summary, our contributions are:
\begin{itemize}
\item We introduce a two-layer meta-evaluation framework for text-to-SVG generation. The controlled
layer, SVGFaith-Diagnostic ($1{,}842$ pairs), pinpoints specific metric blind spots, while the
real-output layer, SVGFaith-Arena ($1{,}327$ pairs), tests reliability on generator outputs without
using an evaluated metric to establish its reference ordering.
\item We expose severe, locatable failures in widely used evaluation methods. In particular, all ten
methods become less accurate when a stronger generator's output is judged less faithful, revealing
a generator-conditioned reliability gap shared even by the strongest methods.
\item We propose SVGFaith-Score, an interpretable pointwise metric whose node scores localise
low-scoring prompt requirements and trace inherited discounts. It leads nine baselines on Arena and
in both Diagnostic directions while using SVGFaith-Decomposer and Qwen3.5-9B at inference. Its
prompt-only graph can be cached across candidates, and all node queries for a rendering can be
scored in one batched, non-autoregressive VQA forward call, making the design compatible with
on-policy reward evaluation.
\end{itemize}

\section{A two-layer benchmark for text-to-SVG faithfulness}
\label{sec:bench}

Both layers of SVGFaith consist of candidate pairs and share one scoring protocol. In each pair, $A$
denotes the candidate that the benchmark labels as more faithful and $B$ the other candidate.
SVGFaith-Diagnostic (\Cref{sec:diagnostic}) has two directions. In the \emph{image-based} direction,
two renderings $I_A$ and $I_B$ share one prompt $x$; $I_A$ satisfies $x$, whereas $I_B$ violates it on
one designated axis. In the \emph{text-based} direction, two descriptions $x_A$ and $x_B$ share one
rendering $I$, which satisfies $x_A$ but contradicts $x_B$ on that axis. In SVGFaith-Arena
(\Cref{sec:arena}), $I_A$ and $I_B$ are two renderings generated from one prompt $x$, and $I_A$ is the
one that human annotators judge more faithful, although it need not satisfy the prompt completely.

Let $S(x,I)$ be a metric's score for text $x$ and rendering $I$. Each candidate is scored on its own,
so no method sees the two candidates together. For Arena and image-based Diagnostic pairs, we form
$d_{\mathrm{image}}$; for text-based Diagnostic pairs, we form $d_{\mathrm{text}}$:

\begin{equation}
d_{\mathrm{image}} = S(x,I_A)-S(x,I_B),
\qquad
d_{\mathrm{text}} = S(x_A,I)-S(x_B,I).
\label{eq:pairwise}
\end{equation}

Let $d$ denote the applicable difference, so a correct decision has $d>0$. Given a threshold
$\tau\ge0$, we count $d>\tau$ as a win, $d<-\tau$ as a loss and $|d|\le\tau$ as a tie.
\emph{Pairwise accuracy} is the fraction of pairs counted as wins. On Arena, it is computed over all
pairs. For each Diagnostic direction, it is computed separately on each of the eight axes and then
averaged with equal weights, so differences in axis size do not affect the direction-level score.
\Cref{sec:exp} fixes $\tau$ and states the remaining reporting conventions.

\subsection{SVGFaith-Diagnostic}
\label{sec:diagnostic}

The image-based direction tests whether a metric resolves a visual difference, and the text-based
direction tests whether it registers a difference in what two descriptions assert. The eight axes are
\emph{negation} (whether forbidden content is absent), \emph{attribute binding} (whether each
property, typically colour, is attached to the correct object), \emph{comparative size} (whether
objects have the specified relative sizes), \emph{spatial relation} (whether objects occupy the
specified relative positions), \emph{object identity} (whether the depicted object matches the
prompt), \emph{counting} (whether the required number of instances is present), \emph{part
completeness} (whether all named parts are present), and \emph{dense long prompt} (whether a local
violation remains detectable within a long compositional description).
Each axis is defined by an objective criterion intended to be judged without appealing to taste. All sixteen axis--direction cells hold at least $100$ pairs (\Cref{tab:diag-axes}), and \Cref{fig:axes-all} shows a contrastive pair from each axis.

\begin{table}[b]
\centering
\caption{Composition of SVGFaith-Diagnostic. Rule-based routes edit a faithful source SVG,
generative routes write both SVGs together, and object identity uses hard-negative retrieval.
Per-axis details are in \Cref{app:details}.}
\label{tab:diag-axes}
\small
\begin{tabular}{llrrr}
\toprule
Axis & Construction route & Image-based & Text-based & Total \\
\midrule
Negation           & Generative & 123 & 123 & 246 \\
Attribute binding  & Generative & 121 & 122 & 243 \\
Comparative size   & Generative & 124 & 124 & 248 \\
Spatial relation   & Generative & 123 & 122 & 245 \\
Object identity    & Hard-negative retrieval & 123 & 118 & 241 \\
Counting           & Instance deletion + generative & 103 & 100 & 203 \\
Part completeness  & Part deletion & 101 & 112 & 213 \\
Dense long prompt  & Single-fact edit & 103 & 100 & 203 \\
\midrule
Total & & 921 & 921 & 1{,}842 \\
\bottomrule
\end{tabular}
\end{table}

\paragraph{Construction.}
We use three construction routes. Rule-based pairs start from a faithful source SVG from SArena or
SVG-Sophia \citep{wang2025internsvg,wang2026reliable} and remove an instance or element, delete named
parts or recolour one named element. Because these pairs edit the vector source directly, all
untargeted vector elements remain unchanged. Object identity instead pairs the faithful rendering with
a visually similar neighbour, so that the substitution cannot be detected from global appearance
alone. Neighbours are retrieved by image similarity and reranked by Qwen3.6-27B
\citep{qwen36_qwen_2026}, which is not among the evaluated metrics. Negation, attribute binding,
comparative size, spatial relation and some counting pairs require coordinated changes across several
elements, so Gemini-3.1-Pro \citep{gemini31pro_google_2026} writes the faithful and violating SVGs
together with their matching prompts. Text-based pairs keep the faithful rendering of the
corresponding image-based construction as $I$. Their violating description $x_B$ either flips the
fact recorded by a rule-based edit or is the text paired with the retrieved or generated violating
rendering.

\paragraph{Quality control.}
Two automatic gates come first. A visibility gate requires the change to be neither invisible at
rendering resolution nor too large for a controlled contrast. A semantic gate, also run with
Qwen3.6-27B, checks whether $A$ is fully faithful, whether $B$ violates the designated axis and
whether the pair is unambiguous. Because this gate is a vision--language model, as are several
evaluated metrics, it only discards pairs and never edits a candidate or decides which candidate is
$A$. Discarding may still bias Diagnostic toward violations that such models can detect. Arena has no
semantic gate, and SVGFaith-Score also leads there (\Cref{sec:exp-main}). Of the $3{,}706$ pairs on the
eight axes that pass both gates, about $2{,}250$ are sampled at random within each axis--direction
cell for human review. Three annotators independently check that $A$ is fully faithful, $B$ violates the designated
axis, the two candidates are distinguishable and the verdict does not depend on subjective taste. We
retain the $1{,}799$ pairs that all three annotators keep. To bring three axis--direction cells to at
least $100$ pairs, $43$ further pairs are checked against the same criteria by expert review rather
than by all three annotators. \Cref{app:details} reports the per-axis operations, the complete funnel
and the rejection reasons.
\label{pg:end3}
\subsection{SVGFaith-Arena}
\label{sec:arena}

SVGFaith-Arena draws its pairs from real text-to-SVG outputs. The prompts are icon and
illustration descriptions from SArena \citep{wang2025internsvg} and a separate set of colour
descriptions. Six systems generate the candidates: three open models (OmniSVG1.1-8B, InternSVG-8B and
SVGen built on Qwen2.5-Coder-7B)
\citep{omnisvg_yang_2025,wang2025internsvg,svgen_wang_2025,qwen25coder_hui_2024}, and three
proprietary models (GPT-5.4, Gemini-3.1-Pro and Claude-Opus-4.6)
\citep{gpt54_openai_2026,gemini31pro_google_2026,claudeopus46_anthropic_2026}. Each pair holds two
outputs for one prompt, from two systems or from two independent samples of one system; the latter
share a style, which isolates differences in content.

Of the $7{,}665$ generated pairs, $6{,}397$ remain after three filters applied in sequence. They
remove pairs with a blank or unrenderable candidate; pairs in which, according to an automatic
Qwen3.6-27B pre-filter \citep{qwen36_qwen_2026}, neither candidate has enough recognisable content
for a faithfulness judgment; and same-system pairs whose candidates are near-duplicates in DINOv2
feature space \citep{oquab2023dinov2}. The pre-filter also places each pair in a preliminary
difficulty bucket, \emph{one-sided} when one candidate appears clearly more faithful and
\emph{high-value} or \emph{uncertain} when the two appear close. The buckets only guide
sampling; all labels come from human judgments. We annotate $2{,}000$ pairs stratified by difficulty
bucket and generator tier (open, proprietary or mixed). Close pairs are over-represented, so Arena is
a challenge set rather than a random sample of the pool (\Cref{app:arena}).

Three annotators independently choose the more faithful candidate or mark a \emph{tie}, seeing the
two in random order without generator identities. The rubric covers prompt faithfulness only:
annotators check objects, counts, colours, spatial relations, parts, negations and composition, and
prefer the candidate that satisfies more of the prompt and violates less of it. Content too
distorted to recognise counts as unsatisfied, and visual polish must not decide the label: a plain
but faithful rendering should not lose to an elaborate off-prompt one.

Labels follow a 2-of-3 majority, and pairs with one vote for each option go to expert adjudication.
Pairwise accuracy requires a winner, so we exclude majority ties and adjudicated pairs that remain
undecided, leaving $1{,}352$ pairs. After running all metrics, we manually review pairs whose
majority label disagrees with several of them, including baselines. On $12$ reviewed pairs, visual
inspection shows that the majority vote points the wrong way, so we correct their labels; we also
remove $25$ invalid or unreliable pairs, yielding the $\mathbf{1{,}327}$ pairs on which all Arena
results are reported.

Across all $2{,}000$ annotated pairs, the pre-adjudication votes ($A$, $B$ or tie) give
Krippendorff's nominal $\alpha=0.568$ \citep{krippendorff_2018} and a raw pairwise agreement of
$0.712$. The $3{,}981$ votes on the $1{,}327$ evaluation pairs give the individual-annotator
reference in \Cref{tab:main}. \Cref{app:arena} reports the label funnel, further agreement
statistics and a sensitivity analysis without adjudicated pairs.
\label{pg:end4}
\section{SVGFaith-Score}
\label{sec:score}

\begin{figure}[t]
\centering
\includegraphics[width=\textwidth]{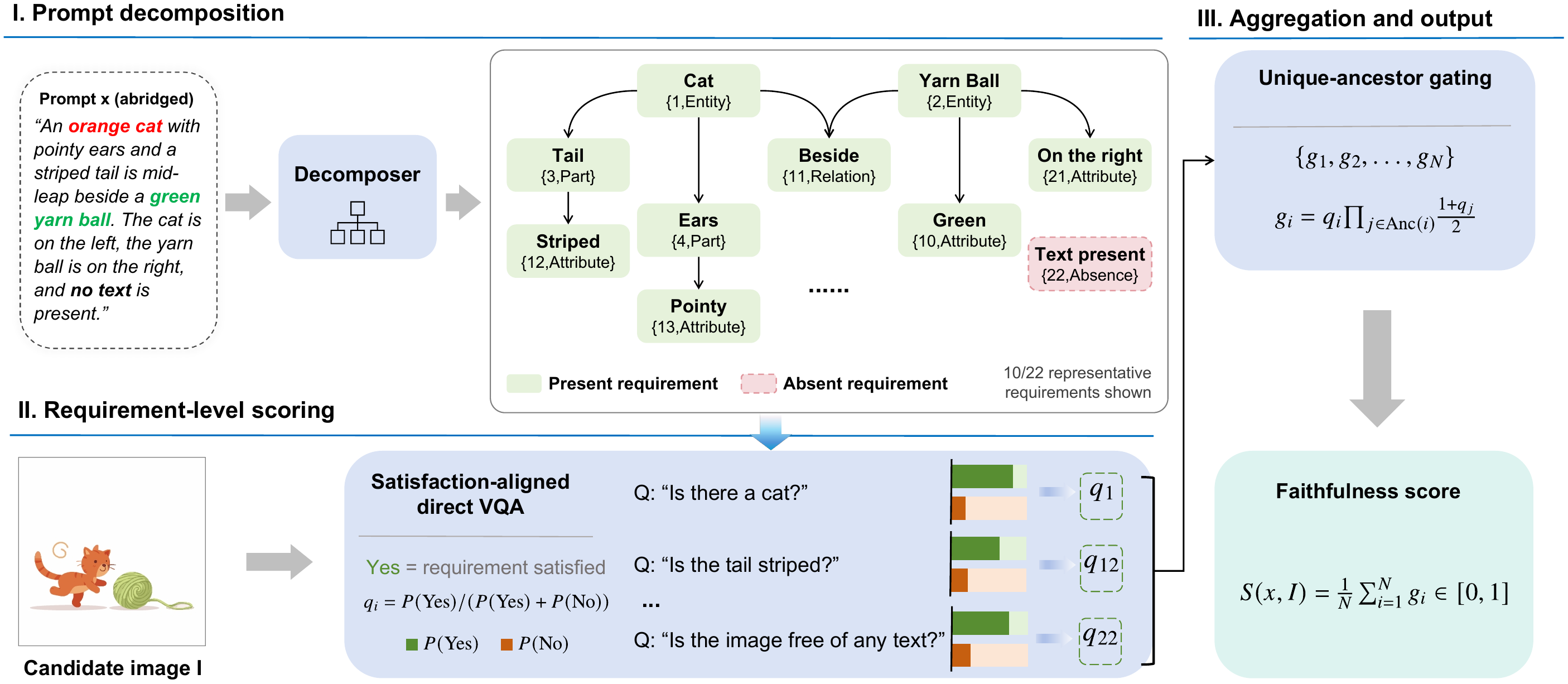}
\caption{SVGFaith-Score. The prompt is decomposed into a graph of atomic requirements without
reference to the image. All requirement queries for a rendering are scored in one batched,
non-autoregressive VQA forward call. Each score is then attenuated by the requirements it
presupposes. The displayed prompt is
abridged. The graph shows ten of the $22$ requirements obtained from the complete prompt.}
\label{fig:overview}
\end{figure}

SVGFaith-Score maps a prompt $x$ and the rasterised rendering $I$ of one candidate to
$S(x,I)\in[0,1]$ in three stages: decomposition into atomic requirements, requirement-level visual
scoring and aggregation.

\subsection{Atomic fact graph}
\label{sec:score-decomp}

SVGFaith-Decomposer, our domain-tuned Qwen3.5-9B \citep{qwen35_qwen_2026}, reads $x$ alone and, after
a reasoning trace, emits a directed acyclic graph whose $N$ nodes $z_1,\ldots,z_N$ are atomic
requirements, each phrased as a yes/no question that can be answered from the candidate's rendered
image alone, in the form introduced by DSG \citep{dsg_cho_2024}.
A node is an \emph{entity} fact, about the presence of an object or part; an \emph{attribute} fact, about the
count, colour, material, shape, size, state or pose of one entity; a \emph{relation} fact, tying
entities by a spatial, comparative or action relation; or a \emph{global} fact, about style, composition or a constraint on the whole image. Each node carries the text span that licenses it, the entities it involves, and whether $x$ requires the fact to hold or forbids it.
An edge $j \rightarrow i$ records that requirement $i$ presupposes $j$, as \emph{the apple is red} presupposes \emph{an apple is present}.

\subsection{Satisfaction-aligned Direct VQA}
\label{sec:score-vqa}

Each requirement $z_i$ is posed directly to a VQA backend as a single yes/no question, with thinking
disabled and no long-form generation. Questions are phrased so that \emph{Yes} always indicates
satisfaction. A presence requirement asks \emph{Are flowers visible?}, whereas an absence
requirement asks \emph{Is the image free of any flowers?} As in VQAScore \citep{vqascore_lin_2024},
we read the answer distribution at the first answer position, but renormalise it over the canonical
\emph{Yes} and \emph{No} tokens:

\begin{equation}
q_i = \frac{P(\mathrm{Yes} \mid I, z_i)}{P(\mathrm{Yes} \mid I, z_i) + P(\mathrm{No} \mid I, z_i)}
\in [0,1],
\label{eq:qi}
\end{equation}

where $P(\cdot \mid I, z_i)$ is the backend's next-token distribution at that position.
Equivalently, $q_i$ is the sigmoid of the \emph{Yes}--\emph{No} logit difference and increases
monotonically with the backend's preference for \emph{Yes}. We use it as a soft satisfaction score.

Because the graph depends only on $x$, it is built once per prompt and reused across all candidate
renderings for that prompt. Given this cached graph, all $N$ requirement queries for one rendering
can be evaluated in one batched, non-autoregressive VQA forward call, and several candidates can
share the same batch. Only the first-answer logits are retained.

\subsection{Unique-ancestor aggregation}
\label{sec:score-agg}

We use the prerequisite graph as a structural prior: a node is discounted when requirements that it
presupposes receive low scores. Let $\mathrm{Anc}(i)$ denote the set of distinct ancestors of node
$i$. For $\alpha\in[0,1]$, we define the soft gate
$h_\alpha(q)=(1-\alpha)+\alpha q$. Because $q\in[0,1]$, the gate lies in
$[1-\alpha,1]$: an ancestor with $q=1$ leaves its descendants unchanged, while a lower-scoring
ancestor discounts them. We then set:

\begin{equation}
g_i = q_i \prod_{j \in \mathrm{Anc}(i)} h_\alpha(q_j),
\qquad
S(x, I) = \frac{1}{N}\sum_{i=1}^{N} g_i,
\label{eq:unique-ancestor}
\end{equation}

with $\alpha=0.5$ fixed across all experiments. At this value, an ancestor scoring $q_j$ scales its
descendants by $(1+q_j)/2$, discounting a low-scoring prerequisite without zeroing the branch.
Setting $\alpha=0$ ignores dependencies and setting $\alpha=1$ with binary $q_j$ recovers the DSG rule
\citep{dsg_cho_2024}, under which a failed prerequisite forces its descendants to zero. Each
distinct ancestor contributes once even if multiple paths connect it to node $i$, avoiding
path-dependent double counting. \Cref{app:method} gives further details.

Besides $S$, we retain $q_i$ and $g_i$. A low $q_i$ marks a requirement the VQA backend scores as
unsatisfied. The gap between $q_i$ and $g_i$ exposes the discount that node $i$ inherits from its
ancestors. Source
text spans and incoming edges map both direct and inherited failures back to the prompt.

\subsection{Model roles and decomposer training}
\label{sec:score-models}

SVGFaith-Score combines SVGFaith-Decomposer and Qwen3.5-9B in separate roles: SVGFaith-Decomposer
converts $x$ into an atomic fact graph, and the released Qwen3.5-9B, used without further training,
scores each requirement against $I$.

We train SVGFaith-Decomposer in three stages organised around the failure modes of graph generation.
Teacher SFT distils reasoning traces and graphs from Qwen3.5-397B-A17B. Rejection-sampling SFT retains
concise, complete and structurally valid student trajectories that match the teacher's fact
signature, falling back to the teacher's trajectory when no sample qualifies.
GRPO \citep{deepseekmath_shao_2024} then orders residual failures through a deterministic hierarchy
from truncation and parsing errors through schema, rule and coverage failures to valid graphs.
A coverage check based on counts of facts, entities, evidence tokens and fact types ranks near-empty
graphs below complete ones, while within-tier criteria penalise repetition and excess length.
Similarity to the teacher's graph is logged for diagnosis and does not enter the reward.
\Cref{app:method} gives the reward, acceptance thresholds and complete training configuration.
\label{pg:end5}
\section{Experiments}
\label{sec:exp}

\subsection{Evaluation setup}
\label{sec:exp-setup}

\paragraph{Protocol.}
Candidates are scored independently and compared via \Cref{eq:pairwise} on Arena ($1{,}327$ pairs) and
each Diagnostic direction ($921$ pairs). Ties receive zero credit, and each Diagnostic score averages
the eight axis accuracies equally. We report $\tau=0$ and $0.01$. For $95\%$ bootstrap intervals, we
resample Arena pairs directly and Diagnostic pairs independently within each axis before equal-weight
aggregation \citep{efron_1979}. For tie analysis, with $W$, $T$ and $L$ the numbers of wins, ties and
losses, we also report half-tie accuracy $(W+0.5T)/(W+T+L)$, conditional direction accuracy
$W/(W+L)$ and non-tie coverage $(W+L)/(W+T+L)$.

\paragraph{Metrics compared.}
We evaluate nine baselines across three families: image--text embeddings (CLIP-L/14, SigLIP-L/16-384, Long-CLIP-L/14, Qwen3-VL-Embedding-8B)
\citep{clip_radford_2021,siglip_zhai_2023,longclip_zhang_2024,qwen3vlembedding_li_2026},
human-preference reward models (PickScore, HPS v2.1, ImageReward)
\citep{pickscore_kirstain_2023,wu2023humanv2,imagereward_xu_2023}, and vision--language scorers (FlanT5-XXL VQAScore and Qwen3.5-122B-A10B used as an LLM-as-a-Judge) \citep{vqascore_lin_2024,qwen35_qwen_2026}.
For the judge, each candidate is evaluated independently from the prompt and a $448{\times}448$ raster, without repeated-score averaging.
The rubric anchors ratings from $1$ (unfaithful) to $5$ (fully faithful) and instructs the model to
ignore aesthetics; we divide the rating by $5$. Initial calls use thinking at temperature $0.5$, and
an unparsable response triggers one constrained recovery call with thinking disabled.
To separate the effect of judge scale, we apply the same protocol to Qwen3.5-9B, and a ten-point
variant of the 9B judge tests whether finer rating categories reduce ties (\Cref{tab:judge-scale}).
The $3{,}981$ individual Arena votes yield a human reference of $0.875$ $([0.865,0.886])$.

\subsection{Comparison with existing metrics}
\label{sec:exp-main}

\begin{table}[t]
\centering
\small
\caption{Pairwise accuracy on Arena and the two Diagnostic directions at $\tau=0$ and
$\tau=0.01$. Diagnostic columns are equal-weight means of the eight per-axis accuracies. The human
row reports mean individual-annotator agreement with the final Arena labels. Complete per-axis
results and the threshold sweep are in
\Cref{tab:axis-image,tab:axis-text,app:sweep}.}
\label{tab:main}
\setlength{\tabcolsep}{4.5pt}
\begin{tabular}{lcccccc}
\toprule
& \multicolumn{2}{c}{Arena} & \multicolumn{2}{c}{Diagnostic (image)}
& \multicolumn{2}{c}{Diagnostic (text)} \\
\cmidrule(lr){2-3}\cmidrule(lr){4-5}\cmidrule(lr){6-7}
Method & $\tau{=}0$ & $\tau{=}0.01$ & $\tau{=}0$ & $\tau{=}0.01$ & $\tau{=}0$ & $\tau{=}0.01$ \\
\midrule
\multicolumn{7}{l}{\emph{Image--text embedding}} \\
CLIP-L/14              & 0.769 & 0.699 & 0.660 & 0.470 & 0.658 & 0.328 \\
SigLIP-L/16-384        & 0.823 & 0.775 & 0.750 & 0.517 & 0.745 & 0.462 \\
Long-CLIP-L/14         & 0.791 & 0.739 & 0.713 & 0.481 & 0.754 & 0.376 \\
Qwen3-VL-Embedding-8B  & 0.861 & 0.843 & 0.777 & 0.717 & 0.799 & 0.668 \\
\midrule
\multicolumn{7}{l}{\emph{Human-preference reward model}} \\
PickScore v1           & 0.833 & 0.605 & 0.799 & 0.313 & 0.737 & 0.149 \\
HPS v2.1               & 0.841 & 0.769 & 0.808 & 0.540 & 0.703 & 0.368 \\
ImageReward v1.0       & 0.824 & 0.819 & 0.784 & 0.779 & 0.782 & 0.767 \\
\midrule
\multicolumn{7}{l}{\emph{Vision--language scoring}} \\
FlanT5-XXL VQAScore    & 0.798 & 0.784 & 0.873 & 0.824 & 0.878 & 0.825 \\
LLM-as-a-Judge (122B)  & 0.754 & 0.754 & 0.902 & 0.902 & 0.857 & 0.857 \\
\midrule
SVGFaith-Score         & \textbf{0.898} & \textbf{0.880} & \textbf{0.965} & \textbf{0.941}
                       & \textbf{0.966} & \textbf{0.953} \\
\midrule
Mean individual annotator & \multicolumn{2}{c}{0.875} & \multicolumn{2}{c}{} & \multicolumn{2}{c}{} \\
\bottomrule
\end{tabular}
\end{table}

\paragraph{Overall accuracy.}
SVGFaith-Score leads all nine baselines on Arena and in both Diagnostic directions at both thresholds
(\Cref{tab:main}). At $\tau=0.01$, it reaches $0.880$ on Arena, $0.941$ on image Diagnostic and
$0.953$ on text Diagnostic (vs.\ baseline bests $0.843$, $0.902$ and $0.857$). All $27$ paired
intervals against the baselines exclude zero at both thresholds, and Arena accuracy closely matches
individual-annotator agreement ($0.875$).

\paragraph{Cross-layer and threshold sensitivity.}
At $\tau=0.01$, Qwen3-VL-Embedding-8B leads baselines on Arena ($0.843$) but drops to $0.717$/$0.668$ on Diagnostic. Scaling the same judge protocol from 9B to 122B-A10B raises Arena/image/text accuracy from $0.655/0.771/0.748$ to $0.754/0.902/0.857$ (\Cref{tab:judge-scale}). SVGFaith-Score remains higher by $0.127/0.039/0.095$, and all three paired intervals exclude zero,
even for the smallest gap on image Diagnostic ($[+0.016,+0.063]$). Moving from $\tau=0$ to $0.01$ reduces SVGFaith-Score by only $0.018/0.024/0.013$, leaving $2.8\%$ of Arena pairs tied and mean per-axis tie rates of $3.7\%/1.7\%$ on image/text Diagnostic. By contrast, PickScore falls from $0.799/0.737$ to $0.313/0.149$ on Diagnostic because its scores occupy a band only a few hundredths wide, so small differences become ties. Complete sweeps are in \Cref{app:sweep}.

\paragraph{Information loss from finite rating categories.}
The 122B judge assigns identical five-point ratings to $17.9\%$ of Arena pairs and reverses only $6.7\%$.
With the 9B judge, the ten-point variant reduces exact ties from $27.1\%$ to $22.7\%$ and raises accuracy from $0.655$ to $0.707$, but still leaves over one pair in five unresolved (\Cref{tab:judge-scale}).
SVGFaith-Score instead aggregates continuous requirement-level probabilities and ties only $2.8\%$ of
Arena pairs at $\tau=0.01$. Giving ties half credit narrows its Arena advantage over the 122B judge from $0.127$ to $0.051$ ($0.894$ vs.\ $0.843$). Conditional on a non-tied decision, the judge reaches
$0.918$ and SVGFaith-Score $0.905$, but their non-tie coverage is $0.821$ and $0.972$. The strict metric therefore reflects both directional correctness and the ability to resolve a pair.

\subsection{Constraint-level and generator-conditioned failures}
\label{sec:exp-slices}

\begin{figure}[t]
\centering
\includegraphics[width=0.9\textwidth]{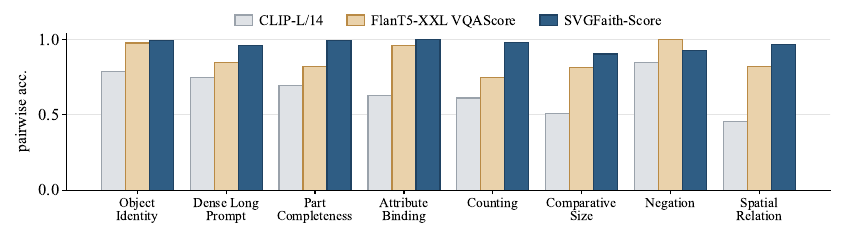}
\caption{Pairwise accuracy at $\tau=0$ for three representative methods on image-based Diagnostic.
The dashed line marks chance. Full results are in \Cref{tab:axis-image,tab:axis-text}.}
\label{fig:axis-chart}
\end{figure}

\paragraph{Constraint-specific performance.}
On image-based Diagnostic at $\tau=0$, CLIP-L/14 reaches only $0.455$ on spatial relations
(\Cref{fig:axis-chart}) and SigLIP-L/16-384 only $0.562$ on attribute binding, while HPS v2.1 ranks
all $103$ counting pairs correctly but reaches only $0.447$ on negation. SVGFaith-Score, by contrast,
remains $\ge0.89$ in all sixteen axis--direction cells at this threshold
(\Cref{tab:axis-image,tab:axis-text}). At $\tau=0.01$, ImageReward ranks second among the baselines
on Arena yet trails FlanT5-XXL VQAScore by $0.045/0.058$ on image/text Diagnostic, so Arena strength
does not ensure constraint discrimination.

\paragraph{Generator-conditioned failures.}
The 122B judge and SVGFaith-Score reach $0.877/0.952$ when Claude-Opus-4.6 is preferred but only
$0.194/0.516$ when it is rejected, with $6/21/4$ and $16/6/9$ wins/ties/losses, respectively. Thus,
the low strict rates do not imply majority reversal: giving ties half credit yields $0.532/0.613$,
and conditional accuracy on non-ties is $0.600/0.640$. The judge similarly records $9/19/8$ when
GPT-5.4 is rejected, while PickScore reaches only $0.349$ when SVGen is preferred
(\Cref{app:polish}). These slices establish a generator-conditioned reliability gap, but their error
types, opponents and difficulty differ, so they do not isolate its cause.

\subsection{Ablations and controls}
\label{sec:exp-ablation}

\begin{table}[t]
\centering
\small
\caption{Component controls on Arena at $\tau=0.01$. $\Delta$ is paired accuracy relative to final
SVGFaith-Score ($95\%$ bootstrap CI); decomposer reuse as VQA is a model-role control.}
\label{tab:component-ablation}
\setlength{\tabcolsep}{2pt}
\begin{tabular}{llccc}
\toprule
Decomposition & VQA backend & Win/tie/loss & Accuracy & $\Delta$ vs.\ final \\
\midrule
None (holistic query)  & Qwen3.5-9B            & $1128/69/130$ & 0.850 & $-0.030$ $([-0.048,-0.012])$ \\
Qwen3.5-9B             & Qwen3.5-9B            & $97/1221/9$   & 0.073 & $-0.807$ $([-0.828,-0.784])$ \\
Qwen3.5-397B-A17B      & Qwen3.5-9B            & $1163/38/126$ & 0.876 & $-0.004$ $([-0.012,+0.005])$ \\
SVGFaith-Decomposer    & SVGFaith-Decomposer   & $1172/30/125$ & 0.883 & $+0.003$ $([-0.010,+0.016])$ \\
SVGFaith-Decomposer    & Qwen3.5-9B            & $1168/37/122$ & 0.880 & $0.000$ \\
\bottomrule
\end{tabular}
\end{table}

\paragraph{Necessity of decomposer domain fine-tuning.}
Replacing SVGFaith-Decomposer with the released Qwen3.5-9B while holding the VQA backend and aggregation fixed reduces Arena accuracy from $0.880$ to $0.073$ ($\Delta=-0.807$ $[-0.828,-0.784]$; \Cref{tab:component-ablation}) and image/text Diagnostic accuracy to $0.063/0.068$, decreases of $0.878$ $[0.857,0.898]$ and $0.885$ $[0.864,0.904]$, respectively.
No question is recovered for $1{,}217/1{,}327$ Arena pairs, $858/921$ image-based pairs and both texts in $818/921$ text-based pairs. These cases exhaust the $40{,}000$-token reasoning budget without a final answer, showing that domain fine-tuning is necessary under this interface.

\paragraph{Decomposition and model-role controls.}
With the same VQA backend, replacing the graph with a single question asking whether the rendering
matches the whole prompt lowers Arena accuracy from $0.880$ to $0.850$ ($\Delta=-0.030$
$[-0.048,-0.012]$; \Cref{tab:component-ablation}), so decomposition improves agreement beyond
holistic scoring. Using the 397B teacher as decomposer gives $0.876$ ($\Delta=-0.004$
$[-0.012,+0.005]$), and reusing SVGFaith-Decomposer as the VQA backend gives $0.883$
($\Delta=+0.003$ $[-0.010,+0.016]$). Neither differs significantly from the final assignment of
SVGFaith-Decomposer to graph construction and the released Qwen3.5-9B to visual scoring.

\section{Conclusion}
\label{sec:conclusion}

This work asks whether an automatic metric can identify the more faithful of two SVG outputs for a
given prompt. SVGFaith-Diagnostic probes this ability under controlled violations along eight
capability axes, varying either the rendering or the description, while SVGFaith-Arena tests
transfer to $1{,}327$ human-annotated comparisons between real generator outputs. Together, the two
layers show which constraints a metric misses and whether it remains reliable when errors co-occur.

On Diagnostic, embedding metrics, human-preference reward models and vision--language scorers all
show constraint-specific weaknesses, in some cases below chance. Their accuracy also decreases on
Arena slices where humans reject outputs from stronger generators. The cause of this
generator-conditioned reliability gap remains open, but the gap shows that aggregate accuracy can
hide systematic failures on important subsets.

SVGFaith-Score addresses fine-grained faithfulness by decomposing a prompt into atomic requirements,
scoring each requirement against one rendering and propagating discounts through prerequisite
relations. At $\tau=0.01$, it reaches $0.880$ on Arena and $0.941/0.953$ on image-/text-based
Diagnostic, outperforming all nine evaluated baselines while exposing interpretable per-requirement
scores. Its drop to $0.516$ when Claude-Opus-4.6 is rejected nevertheless leaves robustness to
generator-conditioned shifts as an open problem. The score also inherits errors from prompt
decomposition and from an uncalibrated VQA readout. Future work should address these error sources
and incorporate native SVG structure rather than relying only on rasterised renderings. Because its
graph is built once per prompt and each candidate is then scored in a single batched forward call,
SVGFaith-Score could also serve as the reward in on-policy reinforcement learning of text-to-SVG
generators, a use we have not yet tested.
More broadly, reporting controlled and real-output evidence separately enables a clearer assessment of progress in faithfulness. A single aggregate benchmark can obscure this distinction.

\label{endofmaintext}
\clearpage

\subsection*{AI use statement}
Generative AI tools were used during paper preparation to improve language, organisation and
presentation; assist literature discovery and citation checking; and draft auxiliary scripts for
verifying reported statistics and formatting figures. The authors checked cited sources, reviewed
AI-assisted code and numerical outputs, and determined all scientific claims and final wording.
The authors take full responsibility for the content of this paper.

\subsection*{Ethics statement}
The benchmarks include human judgements of generated SVG outputs. Annotation was conducted by
internal non-author employees as part of their regular compensated duties, and they were informed
that their judgements would be used for research. The task collected no sensitive personal
information and exposed annotators only to prompts and generated graphics. The activity was
determined to be exempt from formal ethics review. Generator identities were hidden during
annotation, and no annotator identities will be released.

\subsection*{Reproducibility statement}
Benchmark construction and annotation are documented in \Cref{sec:bench} and
Appendices~\ref{app:details}--\ref{app:arena}; the metric and decomposer training are specified in
\Cref{sec:score} and Appendix~\ref{app:method}; and evaluation, bootstrap and control protocols are
reported in \Cref{sec:exp} and Appendix~\ref{app:sweep}. The appendices provide per-axis results,
prompts, model configurations and training hyperparameters. We plan to release the benchmark data,
code and evaluation scripts publicly after the review process.

\bibliography{iclr2027_conference}
\bibliographystyle{iclr2027_conference}

\appendix
\section{Related Work}
\label{sec:related}

\subsection{Text-to-SVG generation}

SVG encodes resolution-independent graphics as structured markup \citep{svg_w3c_1999}; like the
point clouds used in 3D object tracking \citep{focustrack_zhou_2025}, it represents geometry
explicitly rather than on a pixel grid. Early learned approaches modelled sketches, paths or
drawing-command sequences \citep{sketchrnn_david_2018,svgvae_lopes_2019,deepsvg_carlier_2020,clipgen_shen_2022,
im2vec_reddy_2021,strokenuwa_tang_2024}. Text-conditioned synthesis then developed along two main
routes. Optimisation- and diffusion-based methods adjust or predict vector primitives under
raster-space supervision, often through a differentiable rasteriser
\citep{diffvg_li_2020,clipdraw_frans_2022,styleclipdraw_schaldenbrand_2022,
vectorfusion_jain_2023,diffsketcher_xing_2023,nivel_thamizharasan_2024,
t2vecneualpath_zhang_2024,svgdreamer_xing_2024,svgdreamer++_xing_2025,svgfusion_xing_2024}.

Code-centric methods instead emit SVG commands or XML directly
\citep{iconshop_wu_2023,llm4svg_xing_2025,chat2svg_wu_2025,starvector_rodriguez_2025,
omnisvg_yang_2025,wang2025internsvg}, many by fine-tuning large language or vision--language
models; fully quantised training can extend language-model fine-tuning even to edge devices
\citep{gsqtuning_zhou_2025}. Recent systems increasingly incorporate explicit reasoning,
reinforcement learning or rendered visual feedback
\citep{svgen_wang_2025,svgthinker_chen_2025,reasonsvg_xing_2025,rlrf_rodriguez_2025,
wang2026reliable,zhang2025duetsvg,renderintheloop_liang_2026}. These advances improve
renderability, visual quality and controllability, making semantic faithfulness an increasingly
important evaluation target alongside syntax and appearance. We focus on the protocols used to
measure generated outputs rather than proposing another generator.

\paragraph{Adjacent vector-graphics tasks.}
Image-to-vector methods reconstruct editable or layered graphics from raster inputs
\citep{live_ma_2022,samvg_zhu_2024,supersvg_hu_2024,sglive_zhou_2024,
optimizeandreduce_hirschorn_2024,adavec_zhao_2025,layered_wang_2025,
hu2026amodalsvg,layerpeeler_wu_2025,layertracer_song_2025}. Related work studies semantic sketching, fonts,
typography, stylisation and vector customization
\citep{clipasso_vinker_2022,clipascene_vinker_2023,deepvecfont_wang_2021,
wordasimage_iluz_2023,sketchagent_vinker_2025,vectornst_efimova_2023,
vectorpainter_hu_2025,textguided_zhang_2023,stylecustomization_zhang_2025}.
Vector-animation methods further introduce temporal motion and
structure-preservation requirements
\citep{livesketch_gal_2024,flipsketch_bandyopadhyay_2025,groupsketch_liang_2025,
mosketch_liu_2025,vanim_liang_2026}. These adjacent settings motivate native-structure evaluation,
but SVGFaith isolates prompt faithfulness for static text-to-SVG outputs.

\subsection{Benchmarks for vector graphics}

Most vector-graphics benchmarks are designed primarily to compare model capabilities. VGBench
\citep{vgbench_zou_2024} covers understanding and generation across SVG, TikZ and Graphviz;
SVGEditBench and its successor
\citep{svgeditbench_nishina_2024, svgeditbenchv2_nishina_2025} target instruction-based editing;
SVGenius \citep{svgenius_chen_2025} adds complexity stratification over understanding, editing and
generation; UniSVG \citep{unisvg_li_2025} supplies a large unified training and evaluation corpus;
and VectorGym \citep{vectorgym_rodriguez_2026} contributes expert-authored annotations for
sketch-conditioned generation, editing, text-to-SVG and captioning. Generation-specific suites also
include OmniSVG's MMSVGBench \citep{omnisvg_yang_2025} and InternSVG's SArena
\citep{wang2025internsvg}. Their generation tracks use
combinations of reference-based raster metrics, image--text similarities, task-specific checks,
preference scores and holistic vision--language judges. Their primary object of evaluation is the
generator, and the constituent metrics are not systematically stress-tested against controlled
semantic violations.

Recent SVG-specific work instead aligns evaluators with human ratings. SVGauge
\citep{svgauge_zini_2026} combines generated-to-reference visual similarity with semantic similarity
between a generated caption and the input prompt. It is reference-based, whereas both SVGFaith
benchmarks evaluate pointwise prompt-faithfulness scores without supplying a canonical reference
rendering. Concurrent SVG-Score \citep{svgscore_cipriano_2026} adapts CLIP encoders to SVG data and
human preferences and trains a Qwen3-VL judge on ordinal human ratings. SVGFaith differs by
meta-evaluating metrics against independently established contrastive orderings, including controlled
constraint violations and real-output comparisons.

VectorGym additionally validates candidate judges using roughly $674$ ratings from $17$ evaluators
and selects a vision--language model per task by correlation with those ratings
\citep{vectorgym_rodriguez_2026}. This establishes
aggregate human alignment, but does not directly test controlled pairwise discrimination or
attribute an error to a particular prompt constraint. Its rubric intentionally treats clarity and
graphic quality as secondary criteria when semantic match is comparable, reflecting a holistic
quality objective. Our benchmark instead isolates prompt faithfulness and treats generator-associated
visual quality as a potential confound rather than part of the target score (\Cref{app:polish}).

\subsection{Faithfulness metrics for raster generation}

Many automatic scores used in SVG evaluation originate in general image--text modelling or raster
text-to-image evaluation. Embedding similarities compare a prompt and an image in a joint space
\citep{clip_radford_2021, clipscore_hessel_2021, siglip_zhai_2023, longclip_zhang_2024}.
Question-based metrics provide finer-grained alternatives. TIFA \citep{tifa_hu_2023} generates
question--answer pairs and checks them with a VQA model; DSG \citep{dsg_cho_2024} organises atomic,
non-redundant questions into a dependency graph so that a child is not credited when its parent
fails; and VQAScore \citep{vqascore_lin_2024} instead reads the likelihood of \emph{Yes} for one
holistic question. GenEval~2's Soft-TIFA \citep{geneval2_kamath_2025} assigns continuous VQA
probabilities to benchmark-supplied visual atoms and aggregates them with arithmetic or geometric
means. A further family is trained on human preference data
\citep{pickscore_kirstain_2023, wu2023humanv2, imagereward_xu_2023}.

SVGFaith-Score is closest to DSG and Soft-TIFA. Like DSG, it represents dependencies explicitly;
like Soft-TIFA, it retains a graded VQA score for each atomic requirement. It differs from the
reported Soft-TIFA setting by decomposing unrestricted prompts into a dependency graph, and phrases
every query so that \emph{Yes} consistently denotes satisfaction, including for absence
requirements. In \Cref{eq:unique-ancestor}, each distinct ancestor contributes the soft gate
$(1-\alpha)+\alpha q_j$. At $\alpha=1$ with binary answers this reduces to DSG's hard rule, while
the default soft form retains a graded node score. Keeping all $N$ requirements in the denominator
also makes scores directly comparable across images; DSG's \emph{drop} variant instead averages over
an image-dependent subset after removing gated children.

\subsection{Meta-evaluation of metrics}

Metric papers commonly validate against human ratings, as in TIFA and DSG
\citep{tifa_hu_2023,dsg_cho_2024}. Gecko \citep{gecko_wiles_2025} conducts a larger-scale
meta-evaluation with
more than $100$K annotations over four annotation templates and three evaluation tasks, showing
that a metric can lead in one setting and trail in another. T2IScoreScore
\citep{t2iscorescore_saxon_2024} instead organises images into semantic error graphs and tests
whether a metric both orders increasingly erroneous image groups and separates adjacent groups.

CROC \citep{croc_leiter_2026} evaluates both \emph{image-based} and \emph{text-based} comparisons
rather than relying only on correlations.
Five of our eight axes have close counterparts among its eight human-supervised categories:
counting, negation, spatial relation, comparative size and part completeness.

The settings differ in three respects. First, SVG structure permits element-level interventions and
tighter control over non-target content. Second, both CROC
sets are organised around constructed original and contrast prompts, whereas our Arena adds
human-labelled comparisons of real generator outputs, allowing faithfulness and overall visual
quality to vary together. Third, each of our items compares two individual candidates through
independently obtained pointwise scores, whereas CROC$^{\mathrm{syn}}$ uses max-over-samples
aggregation to reduce generation noise.
\section{SVGFaith-Diagnostic construction}
\label{app:details}

\begin{figure}[t]
\centering
\includegraphics[width=\textwidth]{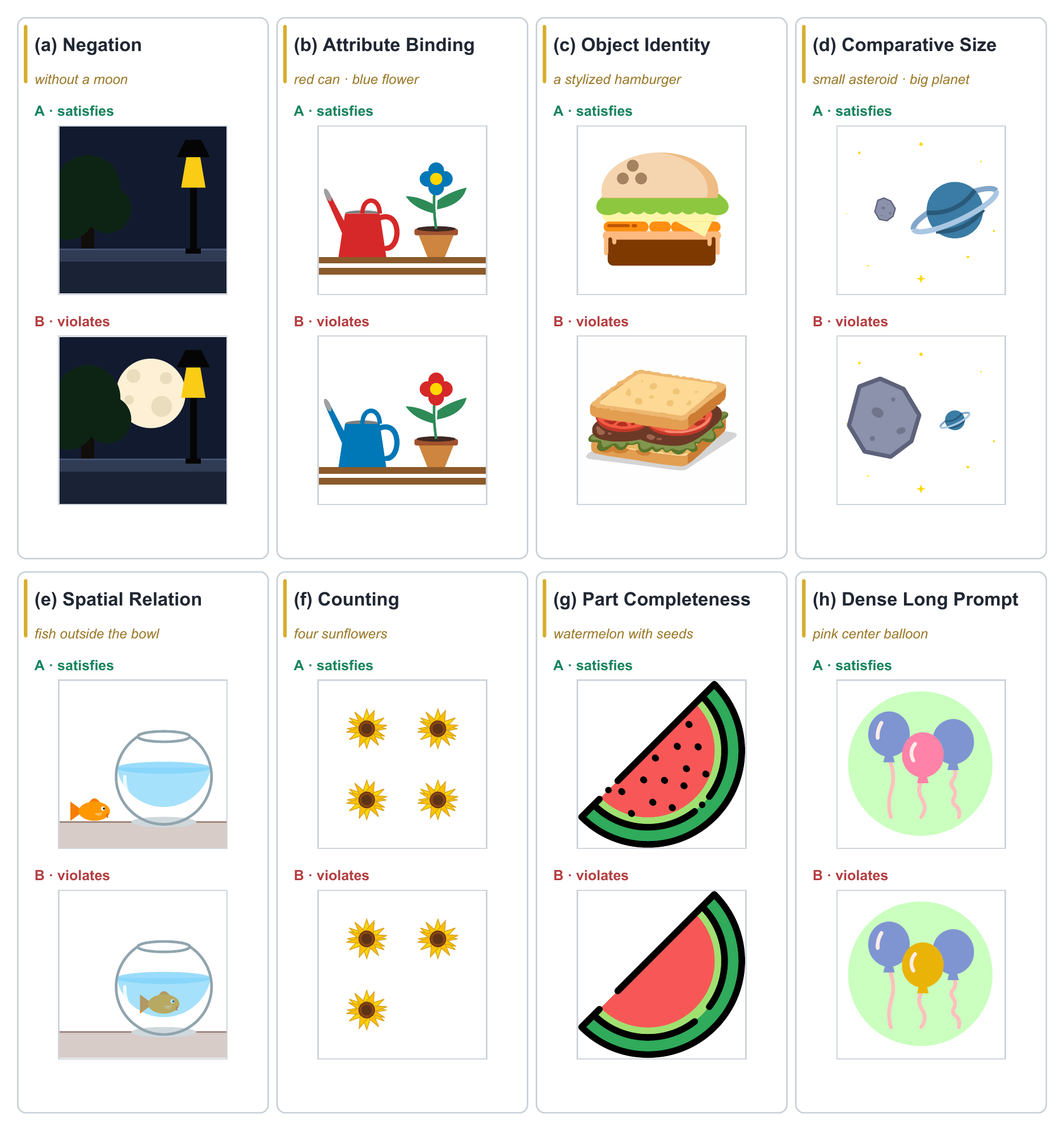}
\caption{One image-based contrastive pair from each capability axis. In every panel, $A$ satisfies
the displayed prompt fragment and $B$ violates the named constraint. Construction routes are
detailed in \Cref{tab:diag-axes}.}
\label{fig:axes-all}
\end{figure}

\Cref{fig:axes-all} shows one image-based contrastive pair from each axis.

\subsection{Sources}

Rule-based edits and identity retrieval start from SVG--prompt pairs in two public corpora:
SArena \citep{wang2025internsvg}, which supplies icons and illustrations with short prompts and
contributes the object identity, counting and part completeness axes, and SVG-Sophia
\citep{wang2026reliable}, whose long, dense descriptions support the dense long prompt axis. Both
are distributed under Apache-2.0. Generatively constructed axes use newly generated SVGs.

\subsection{Rule-based edits and identity retrieval}

For counting, part completeness and dense long prompt, a faithful pair
$(\mathrm{SVG}_A,\text{prompt})$ is transformed into $\mathrm{SVG}_B$ by an edit confined to one
axis:

\begin{itemize}
\item \emph{counting}: one instance is removed from a repeated group, taking $N$ to $N-1$;
\item \emph{part completeness}: one or more parts the prompt names are deleted from the faithful
SVG;
\item \emph{dense long prompt}: one named element is recoloured, or one element or instance is
removed.
\end{itemize}

Object identity uses retrieval rather than source editing. Image--image nearest neighbours of the
faithful rendering are retrieved with SigLIP2-so400m \citep{siglip2_tschannen_2025}, and the top
candidates are reranked by an independent Qwen3.6-27B judge \citep{qwen36_qwen_2026}, which selects a
rendering that is visually close to $A$ while failing the identity stated in the text.

The controlled unit is the semantic axis. A feature a prompt names is often drawn as
a group of paths rather than a single one, so deleting it removes all of them, and a rendering can
lose several visible elements while still differing from $A$ on one axis alone. \Cref{fig:teaser}(a)
is such a case: the eyebrow, the diamond of one eye and the nose are separate paths, and every
difference between the two renderings is a deletion of a named part. An item is discarded when the
edit reaches beyond its axis, or when enough is removed that the violation can no longer be
attributed to the axis it was built for.

\subsection{Generative construction}

Spatial relation, negation, comparative size, attribute binding and some counting pairs are written
by Gemini-3.1-Pro \citep{gemini31pro_google_2026} from a fixed English prompt template; each response
contains the faithful SVG, the violating SVG and the matching pair of prompts. Local geometric
editing is unreliable for these axes: reversing a relation, inverting a size comparison, rebinding an
attribute or inserting a forbidden object requires coordinated changes across several paths. The
independent semantic gate below re-checks every candidate.

\subsection{Text-based items}

The fixed rendering $I$ is the faithful image from the corresponding image-based item. For
rule-based axes, the violating description $x_B$ flips the fact recorded in the edit log. Identity
retrieval uses the neighbour's identity description, and generative axes use the paired prompts
$x_A$ and $x_B$ directly. A candidate $x_B$ that merely omits a detail without asserting anything
false is still faithful and is therefore removed. Generated options may differ in surface form
beyond a single token, but they are retained only when $x_A$ is fully faithful, $x_B$ clearly
violates the designated axis and the ordering remains attributable to that axis.

\subsection{Automatic gates}

The visibility gate measures $\texttt{changed\_fraction}$, the proportion of pixels whose value
changes by more than $30$ in any RGB channel, and requires it to lie in $[0.01, 0.65]$ for
rule-edited or retrieved pairs and $[0.02, 0.65]$ for generated ones. Blank renderings are discarded.
The semantic gate uses Qwen3.6-27B to check whether $A$ is fully faithful, whether $B$
violates the designated axis and whether the pair is discriminative and unambiguous.

\subsection{Human review}

Three annotators independently judged each item selected for review, recording keep or reject and,
when rejecting, one reason from a fixed list: $A$ is not faithful, $B$ does not violate the
designated axis (including the case where it merely omits a detail), both options fail, the two
options are near duplicates, or the item is ambiguous. The $43$ expert-reviewed items fall in three
cells: counting image-based ($24$), part completeness image-based ($6$) and dense long prompt
text-based ($13$). \Cref{tab:diag-funnel} summarises the full funnel; its rule-based rows report
pass rates through the automatic gates rather than a step-by-step account of the final $1{,}842$
pairs.

\begin{table}[t]
\centering
\caption{Construction funnel for SVGFaith-Diagnostic.}
\label{tab:diag-funnel}
\small
\setlength{\tabcolsep}{4pt}
\begin{tabular}{lll}
\toprule
Stage & Scope & Count \\
\midrule
Source pools & SArena icon / illus.; SVG-Sophia & 6{,}013 / 2{,}001 / 49{,}691 \\
Generative, passing visibility gate & spatial relation & 300 $\rightarrow$ 272 \\
 & attribute binding & 160 $\rightarrow$ 153 \\
 & negation & 160 $\rightarrow$ 160 \\
 & comparative size & 160 $\rightarrow$ 156 \\
 & counting & 75 $\rightarrow$ 59 \\
Rule-based, passing automatic gates & dense long prompt & 1{,}000 $\rightarrow$ 148 \\
 & part completeness & 1{,}000 $\rightarrow$ 395 \\
 & counting & 247 $\rightarrow$ 97 \\
After automatic gates & image 2{,}015 + text 1{,}789 & 3{,}804 \\
Eight-axis pool & excluding 98 early colour items & 3{,}706 \\
Presented to annotators & random within each axis--direction cell & $\approx$ 2{,}250 \\
Final & 1{,}799 unanimous + 43 expert & \textbf{1{,}842} \\
\bottomrule
\end{tabular}
\end{table}

\subsection{Image format}

All renderings are opaque three-channel RGB PNGs with no transparent pixels; most are
$384 \times 384$, with a minority of $224 \times 224$ images in the part completeness axis. The two
renderings within a pair always share a resolution.

\section{SVGFaith-Arena details}
\label{app:arena}

\subsection{Filtering}

\begin{table}[h]
\centering
\caption{Automatic filtering of the SVGFaith-Arena pool. Gates are applied in sequence, so each row
acts only on the survivors of the row above.}
\label{tab:arena-funnel}
\small
\begin{tabular}{llrr}
\toprule
Gate & Rule & Removed & Remaining \\
\midrule
& Raw pairs, over 15{,}330 candidate SVGs & & 7{,}665 \\
Render & Either side fails to render or is blank & 282 & 7{,}383 \\
Junk & Qwen3.6-27B marks both candidates unjudgeable & 621 & 6{,}762 \\
Near-duplicate & Same-model pair, DINOv2 CLS cosine $\ge 0.97$ & 365 & 6{,}397 \\
\bottomrule
\end{tabular}
\end{table}

\Cref{tab:arena-funnel} lists the automatic gates applied to the raw pairs. The near-duplicate gate
uses DINOv2 CLS cosine \citep{oquab2023dinov2} and applies only to pairs whose two candidates come
from the same generator.

\subsection{First-round sampling}

We evaluate the first $2{,}000$-pair annotation round. By difficulty bucket, assigned by the
pre-filter as described in \Cref{sec:arena}, it contains $1{,}115$ high-value, $744$ one-sided and
$141$ uncertain pairs: sampling keeps every uncertain pair, over-samples high-value pairs and
under-samples one-sided pairs relative to the pool. By generator tier, $962$ pairs compare two
open-model outputs, $819$ are mixed and $219$ compare two proprietary-model outputs. Pairs of two proprietary outputs and pairs involving
Gemini-3.1-Pro are rare in the pool, so both receive a minimum quota; $134$ sampled pairs involve
Gemini-3.1-Pro. By source, the round contains $783$ icon, $635$ illustration and $582$
colour-description prompts, balanced round-robin within strata defined by difficulty bucket crossed
with generator tier. Sampling uses a fixed seed. The resulting evaluation set is treated as a
deliberately stratified challenge benchmark.

\subsection{From votes to the evaluation set}

\begin{table}[h]
\centering
\caption{From raw votes to the final SVGFaith-Arena evaluation set.}
\label{tab:arena-labels}
\small
\begin{tabular}{lr}
\toprule
Stage & Pairs \\
\midrule
Annotated in the first round & 2{,}000 \\
Three-way disagreement, sent to expert adjudication & 68 \\
\quad of which assigned an expert-adjudicated winner & 40 \\
\quad of which judged indeterminable and left unlabelled & 28 \\
Labelled after adjudication & 1{,}972 \\
\quad of which majority ties, excluded from evaluation & 620 \\
With a decided winner & 1{,}352 \\
Final manual review: winner label corrected & 12 \\
Final manual review: removed as invalid or unreliable & 25 \\
\midrule
Final evaluation set & \textbf{1{,}327} \\
\bottomrule
\end{tabular}
\end{table}

\Cref{tab:arena-labels} traces the annotated pairs to the final evaluation set. The final manual
review took place after all metrics had been run. It examined pairs whose
majority label disagreed with several evaluated metrics, including baselines; pairs on which the
metrics agreed with the majority were not re-examined. The review reversed the winner on $12$
pairs where visual inspection showed that the majority vote pointed the wrong way, and removed
$25$ pairs judged invalid or unreliable. Individual votes were left unchanged, so on the corrected
pairs an annotator who followed the majority counts as incorrect in the individual-annotator
reference.

\subsection{Calibration}

Before annotating, the three annotators jointly labelled a small set of demonstration pairs with
predetermined labels, covering a clear win, a genuine tie, a plain-but-correct rendering against an
ornate-but-off-prompt one, mutually cancelling errors, and one candidate degraded past recognition.
Annotation began once their judgements matched these labels.

\subsection{Agreement}

Coefficients are computed on all $2{,}000$ annotated pairs, from the three individual votes before
adjudication, over the three-way A/B/tie label set; $95\%$ intervals come from a pair-level
bootstrap \citep{efron_1979}. The $28$ pairs left unlabelled after adjudication enter with their original
split votes. Krippendorff's $\alpha$ \citep{krippendorff_2018} is $0.568$ $([0.544, 0.591])$,
Fleiss' $\kappa$ \citep{fleiss_1971} is $0.568$ $([0.544, 0.591])$ and Gwet's AC1
\citep{gwet_2008} is $0.569$ $([0.545, 0.592])$. Raw pairwise agreement is $0.712$
$([0.696, 0.728])$; all three annotators agree on $58.6\%$ of pairs and all three differ on $3.4\%$.

\subsection{Individual-annotator reference on Arena}

Because the individual votes are retained, each annotation can be evaluated against the final label
under the same decision rule as the metrics. A tie vote therefore counts as a loss. Results use the
final $1{,}327$ pairs and all $3{,}981$ individual decisions; the pair-level bootstrap resamples the
three votes on each pair together.

The pooled mean reported in \Cref{tab:main} is $0.875$ $([0.865, 0.886])$; the three annotators
individually reach $0.776$, $0.919$ and $0.931$. This quantifies agreement with the final label,
to which their own votes contribute. A leave-one-annotator-out comparison forms a reference from the
other two where they agree and scores the held-out vote against it, yielding $0.873$
$([0.859, 0.886])$.

\subsection{Sensitivity to the adjudicated pairs}

One of the $40$ expert-adjudicated pairs was removed during final manual review, leaving $39$ in the
evaluation set. \Cref{tab:arena-adjudicated} reports Arena accuracy with and without them. Excluding
them changes each method's accuracy by at most $0.012$ and leaves the ranking of the ten methods
unchanged.

\begin{table}[h]
\centering
\caption{Arena pairwise accuracy at $\tau = 0.01$ with and without the expert-adjudicated pairs.}
\label{tab:arena-adjudicated}
\small
\begin{tabular}{lcc}
\toprule
Method & All ($1{,}327$) & Excluding adjudicated ($1{,}288$) \\
\midrule
CLIP-L/14              & 0.699 & 0.707 \\
SigLIP-L/16-384        & 0.775 & 0.782 \\
Long-CLIP-L/14         & 0.739 & 0.747 \\
Qwen3-VL-Embedding-8B  & 0.843 & 0.852 \\
PickScore v1           & 0.605 & 0.615 \\
HPS v2.1               & 0.769 & 0.776 \\
ImageReward v1.0       & 0.819 & 0.824 \\
FlanT5-XXL VQAScore    & 0.784 & 0.788 \\
LLM-as-a-Judge (122B)  & 0.754 & 0.766 \\
SVGFaith-Score         & \textbf{0.880} & \textbf{0.891} \\
\bottomrule
\end{tabular}
\end{table}

\subsection{Release format}

The left--right placement shown to annotators was randomised per pair and mapped back to the
original candidates at export. In the released files of both benchmarks, the preferred candidate is
normalised into slot $A$. Alongside the faithfulness label used in this paper, the Arena release
carries a secondary preference recorded
on faithfulness-tied pairs, for which annotators were allowed to break the tie on structure and
appearance.

\section{SVGFaith-Score details}
\label{app:method}

\subsection{Why the product runs over distinct ancestors}

The product in \Cref{eq:unique-ancestor} is taken over the set of distinct ancestors of a node
rather than over the paths that reach it. The two differ whenever a node is reachable from the same
prerequisite along more than one route, which happens as soon as two attributes of an entity take
part in a shared relation. A per-path product would then apply that prerequisite's factor more than once, so two graphs
expressing the same requirements would receive different scores if one of them happened to route
the dependency through an extra intermediate node. Aggregating over ancestors makes the score a
function of the requirements rather than of the shape of the derivation.

\subsection{Lenient graph construction at inference}

At inference, decomposer outputs are converted into graphs leniently: every non-empty, answerable
question is kept, recoverable identifier and parent references are repaired, and exact duplicates
are dropped. Strict schema validation is used for filtering training data and measuring
decomposition quality.

\subsection{Teacher data}

New teacher annotations were produced by the same Qwen3.5-397B-A17B model \citep{qwen35_qwen_2026}
through two deployments, one local and one accessed through an API, which contributed $4{,}730$ and
$7{,}512$ accepted examples respectively. These $12{,}242$ new items were merged with $3{,}694$
previously accepted ones. Prompts for the new items come from SVG-Sophia, MMSVG-Illustration and
MMSVG-Icon \citep{wang2026reliable,omnisvg_yang_2025}. Outputs were kept only when the reasoning was
non-empty, the JSON complete and the graph strictly valid; $1{,}963$ new outputs and $152$ older ones
were excluded on those grounds. The resulting $15{,}936$ examples split into $15{,}144$ for training
and $792$ for validation. The $600$ validation examples drawn from the new items, $200$ per source,
were selected by fixed hash without inspecting model scores. Normalised prompt overlap is zero
between training and validation and zero between the new data and both benchmarks.

\subsection{Supervised fine-tuning}

All three stages update every language-model parameter while freezing the vision tower and aligner,
and use ms-swift 4.4.2 \citep{swift_zhao_2025} with FSDP2 \texttt{FULL\_SHARD}
\citep{fsdp_zhao_2023}, BF16 training, TF32, fused AdamW with
$(\beta_1,\beta_2)=(0.9,0.95)$ and gradient clipping at $1.0$. In both SFT stages, loss is computed
only on the final assistant turn, covering the reasoning trace and final graph; packing is
disabled and the maximum sequence length is $32{,}768$. Teacher SFT runs for one epoch on
eight GPUs with one example per rank and an effective batch size of eight: $1{,}893$ optimisation
steps, a peak learning rate of $10^{-5}$, cosine decay, $57$ warmup steps and weight decay $0.1$.

\subsection{Rejection sampling}

Eight samples are drawn for each of $1{,}000$ prompts. A trajectory is accepted only if it is
untruncated, its reasoning is neither empty nor looping, its graph is strictly valid and its
deterministic fact signature matches the teacher's with precision, recall and $F_1$ all at least
$0.95$; the shortest accepted trajectory is kept. Accepted student trajectories exist for $443$
prompts. For the other $557$, where all eight samples failed, the teacher's trajectory is retained,
and the resulting $1{,}000$ examples feed a second round of fine-tuning. This stage also runs for one
epoch on eight GPUs with an effective batch size of eight, giving $125$ optimisation steps; it uses a
peak learning rate of $10^{-6}$, cosine decay, $10$ warmup steps and no weight decay.

\subsection{Reinforcement learning}

GRPO \citep{deepseekmath_shao_2024} starts from the model produced by rejection-sampling SFT. It uses
$316$ distinct training prompts, deterministically padded to $324$ for batching. Each of $18$
rollout rounds processes $18$ prompts and samples eight completions per prompt, or $144$ completions
per generation batch, producing $432$ optimisation steps on six training ranks. Prompt and
completion lengths are capped at $4{,}096$ and $20{,}480$ tokens. Training uses the Dr.~GRPO
objective \citep{drgrpo_liu_2025} with peak learning rate $10^{-6}$, cosine decay, $3\%$ warmup,
$\beta_{\mathrm{KL}}=0.04$ against the frozen rejection-sampling SFT model, clipping $\epsilon=0.2$,
temperature $0.9$, top-$p=0.95$, no reward normalisation and one policy iteration per batch. The
final checkpoint, at step $432$, is used for evaluation.

The reward is staged by output completeness. From lowest to highest, the tiers are truncation
without a usable result, a missing final block, unparseable JSON, a schema violation, a strict-rule
violation, an under-covering graph and a strictly valid graph; within each tier, outputs are ordered
by structural quality, repetition and length. The coverage check inspects fact count, evidence
tokens, entity count and fact types to place near-empty graphs in the under-covering tier.
Teacher-graph string similarity is logged for diagnosis; the online reward is determined entirely by
the staged structural checks.

More formally, let $c(y)$ denote the highest validation tier reached by completion $y$ and let
$u(y)$ denote its within-tier quality. The reward induces the lexicographic ordering
$y_a \succ y_b$ when $c(y_a)>c(y_b)$, or when $c(y_a)=c(y_b)$ and $u(y_a)>u(y_b)$. All eight
completions in a GRPO group answer the same prompt, so the relative advantages compare concrete
failure modes while holding the semantic target fixed.

\section{Additional results}
\label{app:results}

\subsection{LLM-as-a-Judge rating-range and model-scale controls}

\Cref{tab:judge-scale} compares a ten-point Qwen3.5-9B variant with the five-point
Qwen3.5-9B and Qwen3.5-122B-A10B protocols. The two five-point configurations
change only the judge model. Larger model scale improves accuracy on Arena and in
both Diagnostic directions, with the largest gain on image-based Diagnostic.

\begin{table}[h]
\centering
\caption{Pointwise LLM-as-a-Judge controls at $\tau=0.01$. Each cell reports
accuracy/tie rate (\%); ties receive zero credit. Diagnostic columns are
eight-axis equal-weight macro averages.}
\label{tab:judge-scale}
\small
\begin{tabular}{lccc}
\toprule
Judge configuration & Arena & Diagnostic (image) & Diagnostic (text) \\
\midrule
Qwen3.5-9B, ten-point         & 0.707/22.7 & 0.823/17.1 & 0.805/18.3 \\
Qwen3.5-9B, five-point        & 0.655/27.1 & 0.771/22.1 & 0.748/24.3 \\
Qwen3.5-122B-A10B, five-point & 0.754/17.9 & 0.902/\phantom{0}8.6 & 0.857/12.7 \\
\bottomrule
\end{tabular}
\end{table}

For the 9B judge, the ten-point protocol lowers the Arena exact-tie rate by
$4.4$ percentage points and raises accuracy by $0.052$ relative to the
five-point protocol, but still assigns identical scores to $22.7\%$ of pairs.
Within the five-point protocol, scaling from 9B to 122B-A10B raises
Arena/image/text accuracy by $0.099/0.131/0.109$ and reduces the corresponding
tie rates by $9.2/13.5/11.6$ percentage points.

\subsection{Per-axis results}

\begin{table}[h]
\centering
\caption{Per-axis pairwise accuracy for the image-based SVGFaith-Diagnostic. Each cell reports
$\tau=0$ / $\tau=0.01$; axis sample sizes are given in the headers.}
\label{tab:axis-image}
\resizebox{\textwidth}{!}{%
\begin{tabular}{@{}lcccccccc@{}}
\toprule
Method
& \shortstack{Negation\\($n=123$)}
& \shortstack{Attribute\\binding\\($n=121$)}
& \shortstack{Comparative\\size\\($n=124$)}
& \shortstack{Spatial\\relation\\($n=123$)}
& \shortstack{Object\\identity\\($n=123$)}
& \shortstack{Counting\\($n=103$)}
& \shortstack{Part\\completeness\\($n=101$)}
& \shortstack{Dense long\\prompt\\($n=103$)} \\
\midrule
CLIP-L/14                    & 0.846/0.683 & 0.628/0.355 & 0.508/0.379 & 0.455/0.195 & 0.789/0.756 & 0.612/0.291 & 0.693/0.505 & 0.748/0.592 \\
SigLIP-L/16-384              & 0.780/0.675 & 0.562/0.157 & 0.597/0.347 & 0.626/0.195 & 0.976/0.951 & 0.893/0.621 & 0.782/0.574 & 0.786/0.612 \\
Long-CLIP-L/14               & 0.772/0.610 & 0.669/0.223 & 0.468/0.282 & 0.577/0.228 & 0.911/0.878 & 0.641/0.408 & 0.812/0.683 & 0.854/0.534 \\
Qwen3-VL-Embedding-8B        & 0.756/0.707 & 0.661/0.579 & 0.605/0.516 & 0.626/0.512 & 0.927/0.902 & 0.874/0.845 & 0.871/0.832 & 0.893/0.845 \\
\midrule
PickScore v1                 & 0.797/0.293 & 0.636/0.000 & 0.565/0.048 & 0.732/0.073 & 0.886/0.691 & 0.864/0.340 & 0.960/0.624 & 0.951/0.437 \\
HPS v2.1                     & 0.447/0.236 & 0.628/0.050 & 0.734/0.411 & 0.724/0.350 & 0.951/0.837 & 1.000/0.777 & 0.990/0.861 & 0.990/0.796 \\
ImageReward v1.0             & 0.439/0.439 & 0.851/0.843 & 0.694/0.677 & 0.789/0.780 & 0.959/0.959 & 0.748/0.738 & 0.921/0.921 & 0.874/0.874 \\
\midrule
FlanT5-XXL VQAScore          & 1.000/0.984 & 0.959/0.959 & 0.815/0.790 & 0.821/0.715 & 0.976/0.951 & 0.748/0.680 & 0.822/0.772 & 0.845/0.738 \\
LLM-as-a-Judge (122B)        & 0.984/0.984 & 0.983/0.983 & 0.774/0.774 & 0.943/0.943 & 0.992/0.992 & 0.835/0.835 & 0.851/0.851 & 0.854/0.854 \\
\midrule
\textbf{SVGFaith-Score}      & 0.927/0.870 & 1.000/1.000 & 0.903/0.903 & 0.967/0.951 & 0.992/0.992 & 0.981/0.942 & 0.990/0.970 & 0.961/0.903 \\
\bottomrule
\end{tabular}%
}
\end{table}

\begin{table}[h]
\centering
\caption{Per-axis pairwise accuracy for the text-based SVGFaith-Diagnostic, in the same format as
\Cref{tab:axis-image}.}
\label{tab:axis-text}
\resizebox{\textwidth}{!}{%
\begin{tabular}{@{}lcccccccc@{}}
\toprule
Method
& \shortstack{Negation\\($n=123$)}
& \shortstack{Attribute\\binding\\($n=122$)}
& \shortstack{Comparative\\size\\($n=124$)}
& \shortstack{Spatial\\relation\\($n=122$)}
& \shortstack{Object\\identity\\($n=118$)}
& \shortstack{Counting\\($n=100$)}
& \shortstack{Part\\completeness\\($n=112$)}
& \shortstack{Dense long\\prompt\\($n=100$)} \\
\midrule
CLIP-L/14                    & 0.797/0.537 & 0.492/0.107 & 0.661/0.161 & 0.566/0.197 & 0.822/0.712 & 0.660/0.260 & 0.554/0.250 & 0.710/0.400 \\
SigLIP-L/16-384              & 0.902/0.740 & 0.557/0.139 & 0.734/0.250 & 0.533/0.148 & 0.975/0.907 & 0.900/0.670 & 0.679/0.402 & 0.680/0.440 \\
Long-CLIP-L/14               & 0.797/0.504 & 0.541/0.090 & 0.847/0.306 & 0.574/0.221 & 0.907/0.839 & 0.800/0.350 & 0.723/0.339 & 0.840/0.360 \\
Qwen3-VL-Embedding-8B        & 0.935/0.862 & 0.672/0.443 & 0.774/0.597 & 0.582/0.344 & 0.924/0.915 & 0.850/0.700 & 0.795/0.714 & 0.860/0.770 \\
\midrule
PickScore v1                 & 0.878/0.179 & 0.516/0.008 & 0.645/0.016 & 0.500/0.025 & 0.932/0.475 & 0.950/0.280 & 0.768/0.098 & 0.710/0.110 \\
HPS v2.1                     & 0.992/0.854 & 0.533/0.025 & 0.766/0.097 & 0.475/0.139 & 0.958/0.831 & 0.850/0.390 & 0.348/0.098 & 0.700/0.510 \\
ImageReward v1.0             & 0.967/0.943 & 0.934/0.934 & 0.911/0.887 & 0.590/0.574 & 0.983/0.983 & 0.820/0.810 & 0.571/0.554 & 0.480/0.450 \\
\midrule
FlanT5-XXL VQAScore          & 0.967/0.967 & 0.959/0.951 & 0.879/0.831 & 0.844/0.770 & 0.907/0.890 & 0.700/0.590 & 0.946/0.920 & 0.820/0.680 \\
LLM-as-a-Judge (122B)        & 0.878/0.878 & 0.992/0.992 & 0.605/0.605 & 0.893/0.893 & 0.915/0.915 & 0.830/0.830 & 0.964/0.964 & 0.780/0.780 \\
\midrule
\textbf{SVGFaith-Score}      & 1.000/1.000 & 1.000/1.000 & 0.984/0.976 & 0.984/0.984 & 0.975/0.975 & 0.930/0.890 & 0.964/0.938 & 0.890/0.860 \\
\bottomrule
\end{tabular}%
}
\end{table}

\subsection{Generator-conditioned Arena slices}
\label{app:polish}

\begin{figure}[h]
\centering
\includegraphics[width=\textwidth]{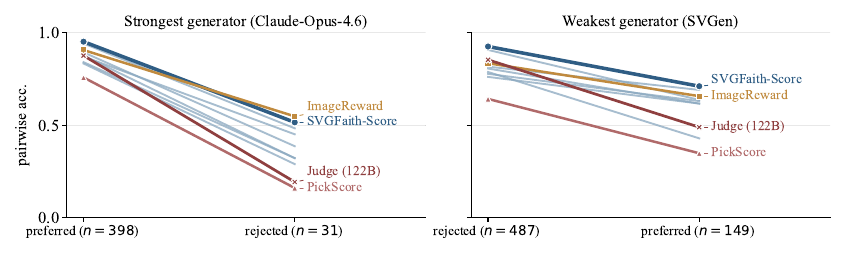}
\caption{Arena pairwise accuracy conditioned on a generator's outcome under the final labels. Each
line is one of the ten methods. \textbf{Left:} pairs won and lost by Claude-Opus-4.6.
\textbf{Right:} pairs lost and won by SVGen, with slices defined as in \Cref{tab:polish}. Every line
falls when the selected generator moves from its more common outcome to its less common one.}
\label{fig:polish}
\end{figure}

\Cref{fig:polish} and \Cref{tab:polish} group Arena pairs by whether a selected generator's output is
the winner or loser under the final label. The rejected slices are smaller for Claude-Opus-4.6 and
GPT-5.4, which win most of their comparisons. As elsewhere, pairwise accuracy is the strict win rate
$W/(W+T+L)$, so ties receive zero credit.

Every method scores lower when Claude-Opus-4.6 or GPT-5.4 is rejected and when SVGen is preferred.
This analysis quantifies a generator-conditioned reliability gap. A strict win rate below $0.5$
does not by itself imply that most pairs are reversed, because ties also receive zero; nor do these
slices isolate a causal mechanism, since their errors, opponents and difficulty can differ.

\begin{table}[h]
\centering
\small
\caption{Arena pairwise accuracy at $\tau=0.01$ on generator-conditioned slices. \emph{Preferred}
contains pairs won by that generator and \emph{rejected} pairs lost by it. For SVGen, the $50$ pairs
between two SVGen outputs are counted only as preferred, so its rejected slice contains only pairs
lost to other generators.}
\label{tab:polish}
\setlength{\tabcolsep}{3.5pt}
\begin{tabular}{lcccccc}
\toprule
& \multicolumn{2}{c}{Claude-Opus-4.6} & \multicolumn{2}{c}{GPT-5.4}
& \multicolumn{2}{c}{SVGen (weakest)} \\
\cmidrule(lr){2-3}\cmidrule(lr){4-5}\cmidrule(lr){6-7}
Method & pref. ($398$) & rej. ($31$) & pref. ($279$) & rej. ($36$)
       & pref. ($149$) & rej. ($487$) \\
\midrule
CLIP-L/14              & 0.834 & 0.290                      & 0.803 & 0.667 & 0.430 & 0.789 \\
SigLIP-L/16-384        & 0.894 & 0.387                      & 0.875 & 0.444 & 0.617 & 0.807 \\
Long-CLIP-L/14         & 0.842 & 0.323                      & 0.849 & 0.500 & 0.617 & 0.762 \\
Qwen3-VL-Embedding-8B  & 0.942 & 0.484                      & 0.932 & 0.694 & 0.638 & 0.908 \\
PickScore v1           & 0.759 & 0.161                      & 0.781 & 0.417 & 0.349 & 0.643 \\
HPS v2.1               & 0.877 & 0.323                      & 0.892 & 0.611 & 0.631 & 0.778 \\
ImageReward v1.0       & 0.907 & 0.548                      & 0.932 & 0.667 & 0.658 & 0.836 \\
FlanT5-XXL VQAScore    & 0.872 & 0.452                      & 0.871 & 0.444 & 0.691 & 0.817 \\
LLM-as-a-Judge (122B)  & 0.877 & 0.194                      & 0.871 & 0.250 & 0.490 & 0.854 \\
\textbf{SVGFaith-Score} & 0.952 & 0.516                     & 0.925 & 0.639 & 0.711 & 0.926 \\
\bottomrule
\end{tabular}
\end{table}

\subsection{Tie-threshold sweep}
\label{app:sweep}

\Cref{tab:sweep} extends the two thresholds of \Cref{tab:main} to seven values of $\tau$, and
\Cref{tab:sweep-tie} reports the corresponding tie rates. SVGFaith-Score leads every baseline in all
six columns of \Cref{tab:main}, and this lead persists on Arena and in both Diagnostic directions
through $\tau=0.02$. With $10{,}000$ paired bootstrap replicates, resampled as in
\Cref{sec:exp-setup}, all $27$ differences against a baseline exclude zero at $\tau=0$; the smallest
is $+0.036$ $([+0.015, +0.057])$ against Qwen3-VL-Embedding-8B on Arena. At $\tau=0.01$, the paired
gaps over the 122B judge are $+0.127$ $([+0.104,+0.150])$ on Arena, $+0.039$ $([+0.016,+0.063])$ on
image Diagnostic and $+0.095$ $([+0.070,+0.121])$ on text Diagnostic.

The sweep also shows that a single absolute $\tau$ is not the same test for every method, which is
why the main table reports both strict ordering ($\tau=0$) and a modest tie band ($\tau=0.01$).
The four embedding metrics, PickScore and HPS all score an image--text pair by the similarity of its
image and text embeddings. These scores occupy bands only a few hundredths wide, so their accuracies
fall steeply as near-zero differences become ties. At $\tau=0.01$, PickScore's mean per-axis tie rate is $68.1\%$ on image Diagnostic; every pair
is tied by $\tau=0.05$, at which point its accuracy is $0.000$ by construction rather than by reversal.
ImageReward, FlanT5-XXL VQAScore and SVGFaith-Score have fewer near-zero pairwise differences and
decay more gently; at $\tau=0.01$, SVGFaith-Score ties only $2.8\%$ of Arena pairs, while its mean
per-axis tie rates are $3.7\%$ and $1.7\%$ on image- and text-based Diagnostic.

The rescaled five-point judge behaves differently: nonzero differences occur in increments of
$0.2$, so its predictions and accuracy remain unchanged throughout the sweep. The $\tau=0.05$ and
$0.1$ columns are therefore not a like-for-like ranking. For methods whose scores lie in $[0,1]$, a
band of $0.1$ is a tenth of the full range: SVGFaith-Score's Arena accuracy falls from $0.898$ to
$0.748$ because its tie rate rises from $0\%$ to $21.3\%$, not because it reverses more pairs
(\Cref{tab:sweep-tie}). ImageReward occupies a wider numerical range, so the same absolute band
leaves most of its pairs decisive and its Arena number stays at $0.785$. The judge is untouched at
every $\tau\le 0.1$. These columns show that a large absolute margin is not a scale-invariant test,
which is why the main comparison uses $\tau=0$ and $0.01$.

\begin{table}[h]
\centering
\caption{Pairwise accuracy over $\tau \in \{0, 0.001, 0.005, 0.01, 0.02, 0.05, 0.1\}$ on Arena and
both Diagnostic directions. Diagnostic values are equal-weight means of the eight per-axis
accuracies. \Cref{tab:main} reports the $\tau = 0$ and $\tau = 0.01$ columns. Bold marks the
highest value in each column.}
\label{tab:sweep}
\footnotesize
\setlength{\tabcolsep}{4.5pt}
\begin{tabular}{lccccccc}
\toprule
Method & $\tau{=}0$ & $0.001$ & $0.005$ & $0.01$ & $0.02$ & $0.05$ & $0.1$ \\
\midrule
\multicolumn{8}{l}{\emph{Arena} ($n = 1{,}327$)} \\
CLIP-L/14              & 0.769 & 0.765 & 0.739 & 0.699 & 0.638 & 0.410 & 0.153 \\
SigLIP-L/16-384        & 0.823 & 0.819 & 0.798 & 0.775 & 0.726 & 0.576 & 0.338 \\
Long-CLIP-L/14         & 0.791 & 0.788 & 0.765 & 0.739 & 0.679 & 0.491 & 0.201 \\
Qwen3-VL-Embedding-8B  & 0.861 & 0.861 & 0.853 & 0.843 & 0.818 & 0.746 & 0.615 \\
PickScore v1           & 0.833 & 0.812 & 0.722 & 0.605 & 0.344 & 0.018 & 0.000 \\
HPS v2.1               & 0.841 & 0.831 & 0.804 & 0.769 & 0.682 & 0.432 & 0.127 \\
ImageReward v1.0       & 0.824 & 0.823 & 0.821 & 0.819 & 0.815 & 0.802 & \textbf{0.785} \\
FlanT5-XXL VQAScore    & 0.798 & 0.793 & 0.789 & 0.784 & 0.764 & 0.718 & 0.656 \\
LLM-as-a-Judge (122B)  & 0.754 & 0.754 & 0.754 & 0.754 & 0.754 & 0.754 & 0.754 \\
SVGFaith-Score         & \textbf{0.898} & \textbf{0.895} & \textbf{0.885} & \textbf{0.880} & \textbf{0.870} & \textbf{0.812} & 0.748 \\
\midrule
\multicolumn{8}{l}{\emph{Diagnostic, image-based} (eight-axis macro)} \\
CLIP-L/14              & 0.660 & 0.645 & 0.577 & 0.470 & 0.331 & 0.119 & 0.029 \\
SigLIP-L/16-384        & 0.750 & 0.729 & 0.619 & 0.517 & 0.393 & 0.182 & 0.049 \\
Long-CLIP-L/14         & 0.713 & 0.686 & 0.581 & 0.481 & 0.318 & 0.125 & 0.039 \\
Qwen3-VL-Embedding-8B  & 0.777 & 0.773 & 0.749 & 0.717 & 0.656 & 0.469 & 0.271 \\
PickScore v1           & 0.799 & 0.717 & 0.486 & 0.313 & 0.113 & 0.000 & 0.000 \\
HPS v2.1               & 0.808 & 0.771 & 0.668 & 0.540 & 0.367 & 0.143 & 0.010 \\
ImageReward v1.0       & 0.784 & 0.784 & 0.783 & 0.779 & 0.767 & 0.745 & 0.700 \\
FlanT5-XXL VQAScore    & 0.873 & 0.859 & 0.845 & 0.824 & 0.779 & 0.688 & 0.574 \\
LLM-as-a-Judge (122B)  & 0.902 & 0.902 & 0.902 & 0.902 & 0.902 & \textbf{0.902} & \textbf{0.902} \\
SVGFaith-Score         & \textbf{0.965} & \textbf{0.960} & \textbf{0.950} & \textbf{0.941} & \textbf{0.921} & 0.832 & 0.640 \\
\midrule
\multicolumn{8}{l}{\emph{Diagnostic, text-based} (eight-axis macro)} \\
CLIP-L/14              & 0.658 & 0.609 & 0.446 & 0.328 & 0.190 & 0.055 & 0.012 \\
SigLIP-L/16-384        & 0.745 & 0.708 & 0.585 & 0.462 & 0.309 & 0.096 & 0.009 \\
Long-CLIP-L/14         & 0.754 & 0.685 & 0.517 & 0.376 & 0.190 & 0.058 & 0.016 \\
Qwen3-VL-Embedding-8B  & 0.799 & 0.789 & 0.721 & 0.668 & 0.523 & 0.289 & 0.121 \\
PickScore v1           & 0.737 & 0.648 & 0.334 & 0.149 & 0.028 & 0.000 & 0.000 \\
HPS v2.1               & 0.703 & 0.650 & 0.489 & 0.368 & 0.235 & 0.039 & 0.001 \\
ImageReward v1.0       & 0.782 & 0.779 & 0.771 & 0.767 & 0.754 & 0.703 & 0.638 \\
FlanT5-XXL VQAScore    & 0.878 & 0.869 & 0.861 & 0.825 & 0.773 & 0.637 & 0.533 \\
LLM-as-a-Judge (122B)  & 0.857 & 0.857 & 0.857 & 0.857 & 0.857 & \textbf{0.857} & \textbf{0.857} \\
SVGFaith-Score         & \textbf{0.966} & \textbf{0.965} & \textbf{0.958} & \textbf{0.953} & \textbf{0.931} & 0.838 & 0.669 \\
\bottomrule
\end{tabular}
\end{table}

\begin{table}[h]
\centering
\caption{Tie rate in percent over the same sweep. Diagnostic values are equal-weight means of the
eight per-axis tie rates. Read alongside \Cref{tab:sweep}: a method whose accuracy falls while its
tie rate rises is losing resolution rather than reversing pairs.}
\label{tab:sweep-tie}
\footnotesize
\setlength{\tabcolsep}{4.5pt}
\begin{tabular}{lccccccc}
\toprule
Method & $\tau{=}0$ & $0.001$ & $0.005$ & $0.01$ & $0.02$ & $0.05$ & $0.1$ \\
\midrule
\multicolumn{8}{l}{\emph{Arena} ($n = 1{,}327$)} \\
CLIP-L/14              & \phantom{0}2.2 & \phantom{0}3.2 & \phantom{0}7.4 & 13.6 & 24.4 & 55.2 & \phantom{0}84.7 \\
SigLIP-L/16-384        & \phantom{0}2.2 & \phantom{0}2.9 & \phantom{0}6.9 & 11.8 & 18.2 & 37.3 & \phantom{0}64.4 \\
Long-CLIP-L/14         & \phantom{0}2.3 & \phantom{0}3.4 & \phantom{0}7.5 & 12.3 & 21.5 & 46.8 & \phantom{0}78.6 \\
Qwen3-VL-Embedding-8B  & \phantom{0}0.0 & \phantom{0}0.5 & \phantom{0}1.8 & \phantom{0}3.5 & \phantom{0}7.4 & 18.8 & \phantom{0}35.6 \\
PickScore v1           & \phantom{0}2.5 & \phantom{0}6.6 & 20.3 & 36.2 & 64.5 & 98.2 & 100.0 \\
HPS v2.1               & \phantom{0}2.2 & \phantom{0}3.6 & \phantom{0}8.4 & 14.2 & 26.8 & 54.7 & \phantom{0}87.3 \\
ImageReward v1.0       & \phantom{0}2.2 & \phantom{0}2.5 & \phantom{0}2.8 & \phantom{0}3.5 & \phantom{0}4.3 & \phantom{0}6.9 & \phantom{00}9.2 \\
FlanT5-XXL VQAScore    & \phantom{0}2.5 & \phantom{0}3.2 & \phantom{0}4.1 & \phantom{0}5.2 & \phantom{0}8.6 & 15.3 & \phantom{0}24.5 \\
LLM-as-a-Judge (122B)  & 17.9 & 17.9 & 17.9 & 17.9 & 17.9 & 17.9 & \phantom{0}17.9 \\
SVGFaith-Score         & \phantom{0}0.0 & \phantom{0}0.4 & \phantom{0}2.2 & \phantom{0}2.8 & \phantom{0}5.2 & 12.7 & \phantom{0}21.3 \\
\midrule
\multicolumn{8}{l}{\emph{Diagnostic, image-based} (eight-axis macro)} \\
CLIP-L/14              & \phantom{0}0.0 & \phantom{0}3.4 & 18.6 & 36.2 & 59.1 & 87.0 & \phantom{0}97.1 \\
SigLIP-L/16-384        & \phantom{0}0.0 & \phantom{0}4.8 & 24.1 & 40.6 & 57.4 & 81.2 & \phantom{0}95.0 \\
Long-CLIP-L/14         & \phantom{0}0.4 & \phantom{0}6.5 & 25.1 & 42.1 & 64.0 & 87.1 & \phantom{0}96.1 \\
Qwen3-VL-Embedding-8B  & \phantom{0}0.0 & \phantom{0}0.7 & \phantom{0}4.9 & 10.4 & 20.2 & 45.0 & \phantom{0}70.6 \\
PickScore v1           & \phantom{0}0.7 & 14.5 & 48.7 & 68.1 & 88.7 & 100.0 & 100.0 \\
HPS v2.1               & \phantom{0}0.0 & \phantom{0}5.1 & 21.6 & 39.7 & 60.6 & 85.5 & \phantom{0}99.0 \\
ImageReward v1.0       & \phantom{0}0.0 & \phantom{0}0.1 & \phantom{0}1.1 & \phantom{0}1.8 & \phantom{0}3.8 & \phantom{0}8.8 & \phantom{0}15.6 \\
FlanT5-XXL VQAScore    & \phantom{0}0.5 & \phantom{0}2.7 & \phantom{0}5.2 & \phantom{0}8.3 & 14.9 & 27.3 & \phantom{0}40.8 \\
LLM-as-a-Judge (122B)  & \phantom{0}8.6 & \phantom{0}8.6 & \phantom{0}8.6 & \phantom{0}8.6 & \phantom{0}8.6 & \phantom{0}8.6 & \phantom{00}8.6 \\
SVGFaith-Score         & \phantom{0}0.0 & \phantom{0}0.8 & \phantom{0}2.5 & \phantom{0}3.7 & \phantom{0}6.3 & 15.8 & \phantom{0}35.7 \\
\midrule
\multicolumn{8}{l}{\emph{Diagnostic, text-based} (eight-axis macro)} \\
CLIP-L/14              & \phantom{0}1.0 & 10.1 & 40.6 & 61.2 & 80.3 & 94.5 & \phantom{0}98.8 \\
SigLIP-L/16-384        & \phantom{0}1.5 & \phantom{0}8.8 & 31.6 & 49.0 & 68.1 & 90.4 & \phantom{0}99.1 \\
Long-CLIP-L/14         & \phantom{0}0.9 & 11.3 & 38.0 & 59.8 & 80.1 & 94.2 & \phantom{0}98.4 \\
Qwen3-VL-Embedding-8B  & \phantom{0}0.0 & \phantom{0}2.1 & 14.1 & 23.6 & 44.1 & 70.6 & \phantom{0}87.9 \\
PickScore v1           & \phantom{0}2.0 & 18.8 & 64.4 & 84.9 & 97.2 & 100.0 & 100.0 \\
HPS v2.1               & \phantom{0}1.0 & 10.6 & 40.1 & 58.5 & 75.6 & 96.1 & \phantom{0}99.9 \\
ImageReward v1.0       & \phantom{0}4.8 & \phantom{0}5.3 & \phantom{0}7.0 & \phantom{0}8.1 & 10.1 & 18.3 & \phantom{0}28.7 \\
FlanT5-XXL VQAScore    & \phantom{0}1.6 & \phantom{0}4.4 & \phantom{0}5.9 & 11.9 & 19.5 & 34.9 & \phantom{0}45.9 \\
LLM-as-a-Judge (122B)  & 12.7 & 12.7 & 12.7 & 12.7 & 12.7 & 12.7 & \phantom{0}12.7 \\
SVGFaith-Score         & \phantom{0}0.0 & \phantom{0}0.1 & \phantom{0}1.1 & \phantom{0}1.7 & \phantom{0}4.6 & 15.1 & \phantom{0}32.6 \\
\bottomrule
\end{tabular}
\end{table}

\end{document}